\documentclass[journal,pagebackref=true,breaklinks=true,colorlinks,bookmarks=false]{IEEEtran}

\usepackage{graphicx}
\usepackage{float}
\usepackage{subfigure}
\usepackage{epsfig}
\usepackage{epstopdf}
\usepackage{indentfirst}
\usepackage{amsfonts}
\usepackage{amsmath,amssymb}
\usepackage{algorithm}
\usepackage{algorithmic}
\usepackage{verbatim}
\usepackage{multirow}
\usepackage{array}
\usepackage{color}
\usepackage{cite}
\usepackage{makecell}
\usepackage{balance}
\usepackage{hyperref}
\usepackage[table]{xcolor}
\usepackage{fancyhdr}
\usepackage{hyperref}
\usepackage{soul}
\usepackage[normalem]{ulem}

\ifCLASSINFOpdf

\else

\fi

\begin{document}

\title{Adaptive Attribute and Structure Subspace Clustering Network}

\author{
    Zhihao~Peng,
    Hui~Liu,
	Yuheng~Jia, ~\IEEEmembership{Member,~IEEE}, 
	Junhui~Hou, ~\IEEEmembership{Senior Member,~IEEE}
\thanks{This work was supported in part by the Hong Kong RGC under Grants CityU 11219019 and 11202320, in part by the Hong Kong University Grants Committee under the Institutional Development Scheme Research Infrastructure Grant UGC/IDS11/19, in part by the National Natural Science Foundation of China under Grant 62106044, in part by the Natural Science Foundation of Jiangsu Province (Grants No BK20210221), and in part by the Jiangsu Provincial Double-Innovation Doctor Program (Grants No JSSCBS20210083) and the ZhiShan Scholar Program of Southeast University. Corresponding authors: \textit{Yuheng Jia; Junhui Hou}.}
\thanks{Z. Peng and J. Hou are with the Department of Computer Science, City University of Hong Kong, Kowloon, Hong Kong (e-mail: zhihapeng3-c@my.cityu.edu.hk; hliu99-c@my.cityu.edu.hk; jh.hou@cityu.edu.hk)}
\thanks{H. Liu is with the School of Computing \& Information Sciences, Caritas Institute of Higher Education, Hong Kong. (e-mail:hliu99-c@my.cityu.edu.hk)}
\thanks{Y. Jia is with the School of Computer Science and Engineering, Southeast University, Nanjing 210096, China, and also with Key Laboratory of Computer Network and Information Integration (Southeast University), Ministry of Education, China (e-mail: yhjia@seu.edu.cn).}
}

\markboth{IEEE Transactions on Image Processing}
{Shell \MakeLowercase{\textit{et al.}}: A Sample Article Using IEEEtran.cls for IEEE Journals}

\maketitle
\makeatletter
\def\ps@IEEEtitlepagestyle{
  \def\@oddfoot{\mycopyrightnotice}
  \def\@oddhead{\hbox{}\@IEEEheaderstyle\leftmark\hfil\thepage}\relax
  \def\@evenhead{\@IEEEheaderstyle\thepage\hfil\leftmark\hbox{}}\relax
  \def\@evenfoot{}
}
\def\mycopyrightnotice{
  \begin{minipage}{\textwidth}
  \centering \scriptsize
  Copyright © 2022 IEEE. Personal use of this material is permitted. However, permission to use this material for any other purposes must be obtained from the IEEE by sending an email to pubs-permissions@ieee.org.
  \end{minipage}
}
\makeatother

\begin{abstract}
Deep self-expressiveness-based subspace clustering methods have demonstrated effectiveness. 
However, existing works only consider the \textit{attribute information} to conduct the self-expressiveness, limiting the clustering performance. 
In this paper, we propose a novel adaptive attribute and structure subspace clustering network (AASSC-Net) to simultaneously consider the \textit{attribute and structure information} in an adaptive graph fusion manner. 
Specifically, we first exploit an auto-encoder to represent input data samples with latent features for the construction of an attribute matrix. 
We also construct a mixed signed and symmetric structure matrix to capture the local geometric structure underlying data samples. 
Then, we perform self-expressiveness on the constructed attribute and structure matrices to learn their affinity graphs separately. 
Finally, we design a novel attention-based fusion module to adaptively leverage these two affinity graphs to construct a more discriminative affinity graph. 
Extensive experimental results on commonly used benchmark datasets demonstrate that our AASSC-Net significantly outperforms state-of-the-art methods. 
In addition, we conduct comprehensive ablation studies to discuss the effectiveness of the designed modules. 
The code is publicly available at \url{https://github.com/ZhihaoPENG-CityU/AASSC-Net}.
\end{abstract}


\IEEEpeerreviewmaketitle

\section{Introduction}
Clustering, which aims to partition samples into multiple groups, is broadly employed in many real-world applications, such as image recognition \cite{zhang2017robust,zhang2019ae2,Park_2021_CVPR,jia2021clustering,li2021contrastive}, object segmentation \cite{li2020superpixel,jia2020constrained,huang2020deep,Cho_2021_CVPR}, and text clustering \cite{zhang2017latent,zhang2018generalized,peng2021maximum,jia2021multi}. 
In the last decade, numerous researchers have paid attention to the self-expressiveness (SE)-based subspace clustering, which includes two main steps. First, the SE priori is exploited to learn an affinity graph by assuming each sample can be expressed as a linear combination of other samples within the same subspace. 
Then, spectral clustering \cite{ng2002spectral} is applied to the resulting affinity graph to seek the clustering results. Obviously, learning a clustering-friendly affinity graph plays a critical role in clustering, and accordingly, a family of approaches based on different kinds of regularization were proposed. For example, Elhamifar \emph{et al.} \cite{elhamifar2013sparse} employed the sparse regularization to find a nontrivial sparse representation. Ji \emph{et al.} \cite{ji2014efficient} designed the Tikhonov regularization to yield dense connections within each cluster. Liu \emph{et al.} \cite{liu2012robust} developed the low-rank regularization to explore the lowest rank representation. You \emph{et al.} \cite{you2016oracle} further designed the elastic net regularization to balance the subspace-preserving (sparse solutions) and connectivity (dense solutions). Although these works have obtained impressive clustering performance, one primary limitation is that they cannot handle data with complex nonlinearity well. To this end, some kernel extensions of the existing subspace clustering algorithms \cite{patel2014kernel,yin2016kernel,xia2020nonconvex} were investigated to explore the nonlinearity of data. Unfortunately, the clustering performance of these methods heavily relies on the choice of the kernel function, while a confident rule for selecting the kernel function is still absent in practice. 

\begin{figure}[]
	\centering
	\subfigure[Previous deep SE-based methods]{
	\includegraphics [width=0.99\columnwidth]{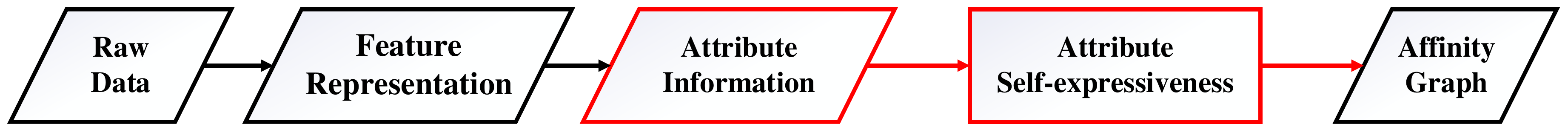}
	}
	\subfigure[Our method]{
	\includegraphics [width=0.99\columnwidth]{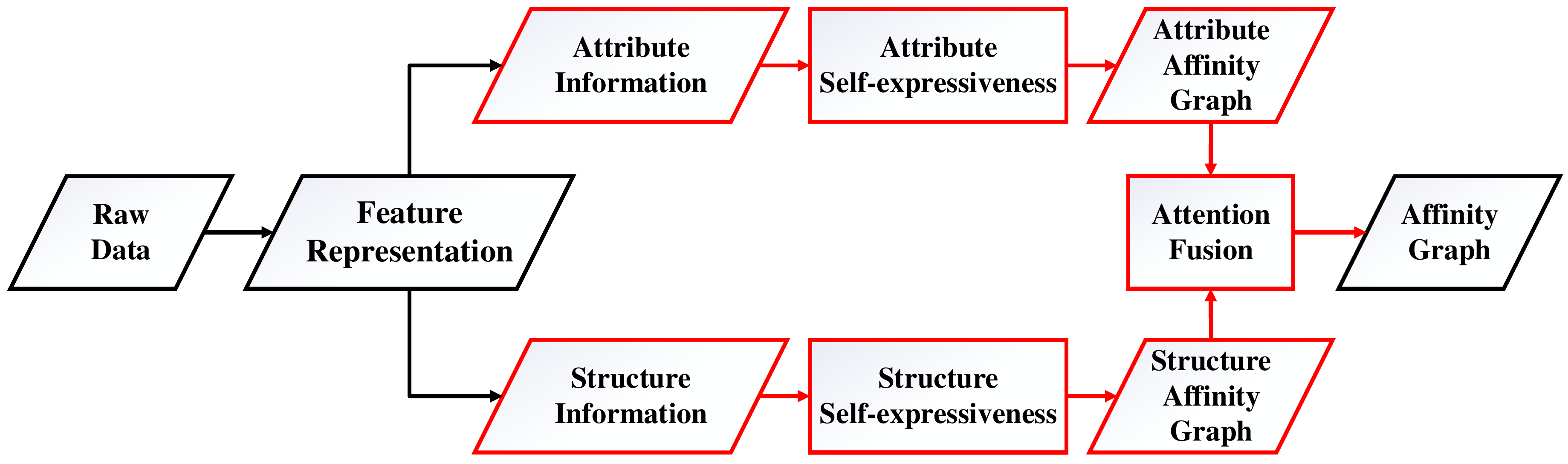}
	}
	\caption{The flowcharts of (a) previous deep SE-based methods and (b) the proposed method. Here, the `\emph{attribute information}' refers to the feature representations of samples in a latent space, and the `\emph{structure information}' to the similarity relationship of a typical sample with the other samples.}
	\label{fig: flowchart}
\end{figure}

Benefiting from the powerful representation ability of deep learning, deep SE-based subspace clustering has a tremendous step, effectively handling the issues mentioned above. 
For example, 
Hinton \emph{et al.} \cite{hinton2006reducing} designed the auto-encoder (AE) method to map the raw data to a nonlinear latent space, in which the clustering results can be obtained by imposing a traditional subspace clustering (e.g., sparse subspace clustering \cite{elhamifar2009sparse}) on the latent feature representations. 
Ji \emph{et al.} \cite{ji2017deep} developed a deep subspace clustering (DSC) framework by inserting a fully-connected layer between the encoder and the decoder to mimic the SE priori. 
Afterward, numerous DSC-based works were proposed by introducing additional regularization terms, such as adversarial learning \cite{zhou2018deep} and pseudo-supervision \cite{lv2021pseudo}. 
All these deep SE-based approaches construct an affinity graph for clustering by performing the SE on the attribute information of latent feature representations. 
However, we argue that the geometric structure information underlying data samples is also valuable for clustering, i.e., for two samples belonging to an identical subspace, their similarity relationships with the other samples are typically similar \cite{vidal2016generalized}. Thus, it is expected that the clustering performance can be boosted by exploring such geometric structure information. 

In this paper, we propose a novel deep SE-based subspace clustering framework to simultaneously consider the attribute and structure information of data in an adaptive graph fusion manner, as shown in Figure \ref{fig: flowchart} (b). Specifically, we first adopt an AE to extract a latent feature representation as the attribute matrix. 
We also exploit the extracted latent representation to construct a structure matrix to capture the local geometric structure underlying data samples. 
Since the underlying geometric structure of a sample is generally similar to those of other samples within the same subspace, our method performs the SE on not only the attribute matrix but also the structure matrix to learn their affinity graphs. Precisely, we achieve these two SE-based reconstructions via two fully-connected layers without the bias and activation function. 
Finally, we propose an attention-based fusion module to adaptively leverage those two affinity graphs to construct a more discriminative one. We validate the superiority of the proposed method over state-of-the-art methods by conducting quantitative and qualitative results on five commonly used benchmark datasets. In addition, to better understand the proposed network, we conduct comprehensive ablation studies of the designed modules. 

The rest of this paper is organized as follows. We review some related works in Section \ref{sec: related work} and introduce our method in Section \ref{sec: proposed method}. Section \ref{sec: eprm} presents the experimental results and analyses. Finally, we conclude this paper in Section \ref{sec: conclusion}.

\section{Related Work}
\label{sec: related work}

We summarize the main notations in Table \ref{tab: notation}, where operators are denoted by calligraphy ones, matrices by bold upper case ones, vectors by bold lower case letters, and scalars by italic lower case letters, respectively. Given a matrix $\mathbf{A}\in\mathbb{R}^{m\times n}$, $ \left\|\mathbf{A}\right\|_F$ denotes the Frobenius norm, i.e., $\left\|\mathbf{A}\right\|_F=\sqrt{\sum_{i=1}^{m}\sum_{j=1}^{n}{\left| a_{i,j} \right|}^2}$.

\begin{table}[t]
  \caption{Summary of notations used in this paper.}
  \label{tab: notation}
  \centering
    \begin{tabular}{l|l}
    \hline\hline
    Notation          & Description                     \\   
    \hline\hline
    $\mathbf{X}\in\mathbb{R}^{\hat{d} \times n}$        & The raw data                      \\
    $\hat{\mathbf{X}}\in\mathbb{R}^{\hat{d} \times n}$     & The reconstructed data   \\
    $\hat{\mathbf{Z}}\in\mathbb{R}^{p_1\times p_2\times p_3\times n}$        & The extracted feature      \\
    $\mathbf{Z}\in\mathbb{R}^{d\times n}$        & The matrix form of $\hat{\mathbf{Z}}$      \\
    $\mathbf{Z}^\mathsf{T}\in\mathbb{R}^{n\times d}$  & The matrix transpose of $\mathbf{Z}$ \\
    $\mathbf{Z}_\emph{A}\in\mathbb{R}^{d\times n}$    & The attribute matrix                   \\ 
    $\mathbf{Z}_\emph{S}\in\mathbb{R}^{n\times n}$    & The structure matrix                    \\ 
    $\mathbf{C}_\emph{A}\in\mathbb{R}^{n\times n}$    & The learned affinity matrix of $\mathbf{Z}_\emph{A}$ \\ 
    $\mathbf{C}_\emph{S}\in\mathbb{R}^{n\times n}$    & The learned affinity matrix of $\mathbf{Z}_\emph{S}$ \\
    $\mathbf{C}_\emph{F}\in\mathbb{R}^{n\times n}$    & The fused affinity matrix               \\
    $\mathbf{I}$                                      & The identity matrix \\
    $\Theta_E$         & The encoder parameters              \\
    $\Theta_D$         & The decoder parameters              \\
    $n$             & The number of samples                 \\
    $\hat{d}$       & The dimension of $\mathbf{X}$           \\
    $p_1\times p_2\times p_3$  & The dimension of $\hat{\mathbf{Z}}$        \\
    $d$             & The dimension of $\mathbf{Z}$           \\
    \hline
    $\left|\cdot\right|$        & The absolute value operation                          \\
    $ \left\|\cdot\right\|_F$   & The Frobenius norm of a matrix                                  \\
    $\emph{f}_\emph{E}$     & The nonlinear mapping function of the encoder    \\
    $\emph{f}_\emph{D}$     & The nonlinear mapping function of the decoder    \\
    $\emph{f}_\emph{AE}$    & The nonlinear mapping function of the AE             \\
    $\emph{f}_\emph{AF}$    & The attention-based fusion module to obtain $\mathbf{C}_\emph{F}$  \\
    \hline\hline
    \end{tabular}
\end{table}

\begin{figure*}[t]
	\centering
	\includegraphics [width=2.02\columnwidth]{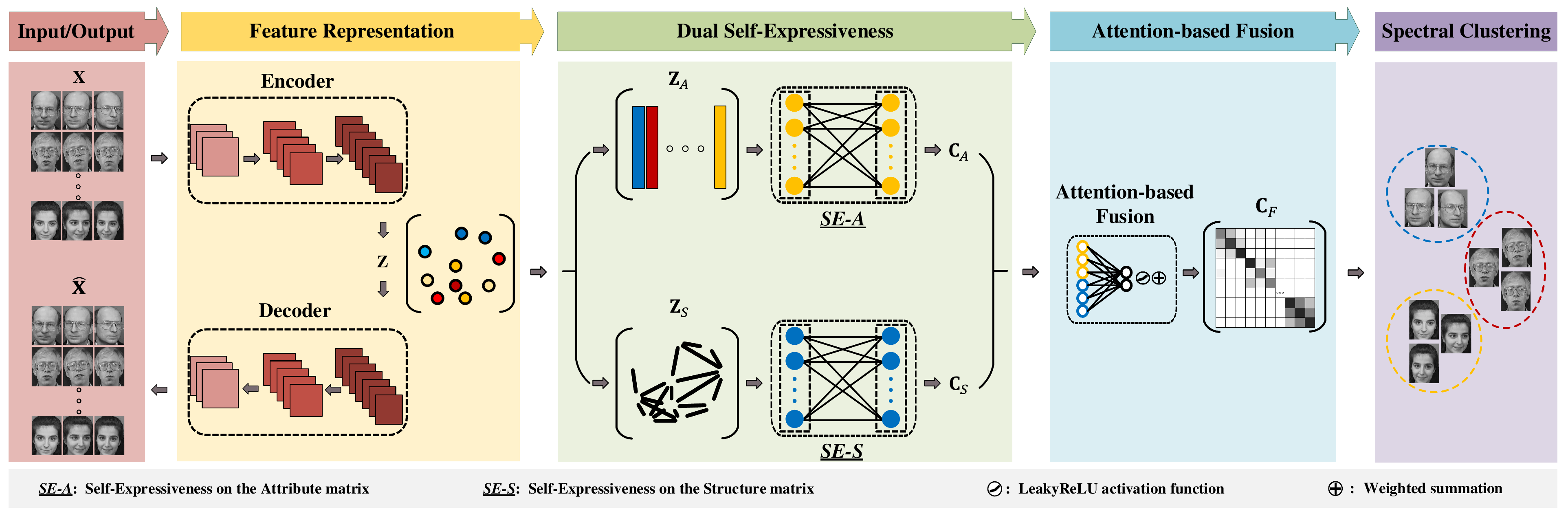}
	\caption{Illustration of the proposed framework, consisting of a feature representation module, a dual self-expressiveness (SE)-based module, and an attention-based fusion module. The feature representation module composed of a series of convolutional layers first generates a nonlinear low-dimensional representation for the raw data, then the dual SE-based module constructed with two fully-connected layers without the bias and activation function learns affinity graphs by exploring both attribute and structure information of data samples, and the attention-based fusion module with a multilayer perceptron finally fuses those two affinity graphs adaptively by performing a weighted sum operation. The clustering results can be obtained by performing spectral clustering on the fused affinity graph.}
	\label{fig: Our_framework}
\end{figure*}

\subsection{Traditional SE-based Methods}

Let $\mathbf{X} = [\mathbf{x}_1,...,\mathbf{x}_{n}]\in\mathbb{R}^{\hat{d} \times n}$ be the raw data with $\hat{d}$ and $n$ being the dimension and number of samples, respectively. We can generally formulate the traditional SE-based models as
\begin{equation}
\begin{aligned}
& \min_{\mathbf{C}} f\left(\mathbf{C}\right) \quad\rm{s.t.}\quad \mathbf{X} = \mathbf{X}\mathbf{C}, \quad diag(\mathbf{C})=\mathbf{0},
\label{eq:SC}
\end{aligned}
\end{equation}
where $\mathbf{C}\in\mathbb{R}^{n\times n}$ is a learned affinity matrix, $diag(\mathbf{C})=\mathbf{0}$ is used to prevent the trivial solution $\mathbf{C}=\mathbf{I}$ with $\mathbf{I}$ being the identity matrix, and $\emph{f}\left(\mathbf{C}\right)$ is used to regularize $\mathbf{C}$ to promote a unique and meaningful solution. By imposing different kinds of regularization on $\mathbf{C}$, a family of works were proposed \cite{elhamifar2013sparse,ji2014efficient,you2016oracle,li2018subspace,you2019affine,liu2019robust,zhang2021robust,wei2021subspace,tao2020latent,chen2021generalized,chen2021low}. 
For example, 
Elhamifar \emph{et al.} \cite{elhamifar2013sparse} exploited a sparse subspace clustering (SSC) method to find a nontrivial sparse representation for $\mathbf{C}$. You \emph{et al.} \cite{you2016scalable} improved SSC by introducing an orthogonal matching pursuit. Ji \emph{et al.} \cite{ji2014efficient} proposed an efficient dense subspace clustering (EDSC) method to yield dense connections within each cluster. Liu \emph{et al.} \cite{liu2012robust} developed a low-rank representation (LRR) method to seek the lowest rank representation among samples. \textcolor{black}{You \emph{et al.} \cite{you2016oracle} designed an elastic net subspace clustering (ENSC) to balance the subspace-preserving (sparse solutions) and connectivity (dense solutions) by leveraging the geometric structure.} Although these methods can produce a meaningful $\mathbf{C}$, they only investigate the linear relationship among samples. To this end, Patel \emph{et al.} \cite{patel2014kernel} explored the nonlinear relationship in the input space by kernel sparse subspace clustering (KSSC). However, determining the kernel function with the associated hyperparameters is generally challenging \cite{kang2017twin}.

\subsection{Deep SE-based Methods}
Deep SE-based methods \cite{hinton2006reducing,butepage2017deep,yao2019deep,zhong2020nonnegative,zhang2021learning} aim to achieve clustering by exploiting the deep nonlinear representation ability. For example, Hinton \emph{et al.} \cite{hinton2006reducing} proposed the auto-encoder (AE) method to extract a latent feature representation, in which the clustering can be achieved by imposing SSC on the latent feature representation. Ji \emph{et al.} \cite{ji2017deep} designed a deep subspace clustering network framework (DSC-Net) to mimic the SE priori and developed the DSC-Net-L1 method with the sparse regularization and the DSC-Net-L2 method with the Tikhonov regularization. Zhou \emph{et al.} \cite{zhou2018deep} proposed deep adversarial subspace clustering (DASC) to supervise the deep representation learning with adversarial learning. Lv \emph{et al.} \cite{lv2021pseudo} used a pseudo-supervised deep subspace clustering (PSSC) approach to weigh the AE reconstruction by integrating the local structure-preserving, SE rule, and pseudo-supervision. However, these previous deep SE-based approaches only focus on the attribute information and barely the geometric structure information to conduct the SE, which inevitably depresses the discriminative ability of the learned affinity graph, as both those two kinds of information are of great importance.

\textcolor{black}{In this paper, we propose an adaptive attribute and structure subspace clustering network to learn a discriminative affinity graph, which simultaneously models the attribute information and the structure information via the self-expressiveness module and subsequently exploits the attention-based fusion module to integrate the learned self-expressiveness graphs.}

\section{Proposed Method}
\label{sec: proposed method}
As shown in Figure \ref{fig: Our_framework}, the proposed framework consists of three key modules, i.e., a feature representation module, a dual self-expressiveness (SE)-based module, and an attention-based fusion module, which will be detailed one by one as follows.

\subsection{Feature Representation Module}
We use an auto-encoder (AE) composed of a series of encoders and decoders to project the raw data $\mathbf{X}$ into a nonlinear latent feature space. Let $\Theta_E$ and $\Theta_D$ be the encoder and decoder parameters, respectively, the optimization problem of the AE can be written as
\begin{equation}
\begin{aligned}
& \min_{\mathbf{Z}} \frac{1}{2} \left\| \mathbf{X} - \hat{\mathbf{X}} \right\|^2_F \quad \rm{s.t.} \quad \mathbf{Z} =\emph{f}_\emph{AE}\left( \mathbf{X}\right),
\label{eq: AE}
\end{aligned}
\end{equation}
where $\hat{\mathbf{Z}}\in\mathbb{R}^{p_1\times p_2\times p_3 \times n}$ is the extracted latent feature representation that reshaped as a data matrix $\mathbf{Z}\in\mathbb{R}^{d \times n}$, $\hat{\mathbf{X}}=f_D(\hat{\mathbf{Z}};\Theta_D)$ is the output of the decoder, $\emph{f}_\emph{AE}\left(\cdot\right)$ denotes the nonlinear mapping function of the AE, and $\left\| \mathbf{X} - \hat{\mathbf{X}} \right\|^2_F$ measures the AE-based reconstruction loss between the input $\mathbf{X}$ and the output $\hat{\mathbf{X}}$.

\subsection{Dual SE-based Module}
As aforementioned, previous deep SE-based methods only explore the attribute information of $\mathbf{Z}$, while neglecting the underlying geometric structure information, which is also of great importance, and thus, the discriminative ability of the learned affinity graph is still limited. To this end, we propose the dual SE-based module to simultaneously explore both the attribute and structure information of $\mathbf{Z}$.

\subsubsection{Attribute SE Formulation} 
To learn an affinity graph from the attribute information, we use a fully-connected layer without the bias and activation function to mimic the SE process. By imposing the Frobenius norm, the attribute SE formulation can be written as
\begin{equation}
\begin{aligned}
& \min_{\mathbf{C}_\emph{A}}
\left\| \mathbf{C}_\emph{A} \right\|^2_F 
+ \gamma \left\| \mathbf{Z}_\emph{A} - \mathbf{Z}_\emph{A}\mathbf{C}_\emph{A} \right\|^2_F,
\label{eq: attribute_Frob}
\end{aligned}
\end{equation}
where $\mathbf{Z}_\emph{A}$ is the attribute matrix, $\mathbf{C}_\emph{A}\in\mathbb{R}^{n\times n}$ is the learned affinity graph from $\mathbf{Z}_\emph{A}$, $\left\| \mathbf{Z}_\emph{A} - \mathbf{Z}_\emph{A}\mathbf{C}_\emph{A} \right\|^2_F$ measures the attribute SE loss. Here, as the AE already owns excellent low-dimensional feature representation ability, we set 
$\mathbf{Z}_\emph{A}=\mathbf{Z}$. In addition, we drop the constraint $diag(\mathbf{C}_\emph{A})=\mathbf{0}$ since here we minimize the Frobenius norm of $\mathbf{C}_\emph{A}$, and the identity matrix $\mathbf{I}$ has large values of the Frobenius norm. That is, the optimization problem in Eq. (\ref{eq: attribute_Frob}) automatically prevents the trivial solution $\mathbf{C}_\emph{A}=\mathbf{I}$.

\subsubsection{Structure SE Formulation}
\label{subsec: s-se}
As $\mathbf{C}_\emph{A}$ obtained from Eq. (\ref{eq: attribute_Frob}) naturally encodes the relationships among samples, we empirically construct a structure matrix $\mathbf{Z}_\emph{S}$ for capturing the underlying geometric structure information of $\mathbf{Z}$, i.e.,
\begin{equation}
    \begin{aligned}
    \mathbf{Z}_\emph{S}
    =\frac{\mathbf{C}_\emph{A}+\mathbf{C}_\emph{A}^\mathsf{T}}{2},
    \label{eq: our_ze}
    \end{aligned}
\end{equation}
where the reason for the symmetrization strategy is to make sure that nodes $i$ and $j$ get connected to each other if either $\mathbf{x}_i$ or $\mathbf{x}_j$ is in the representation of the other, the reason for the mixed-signs strategy is that the negative values in the coefficient matrix include the discriminative information among samples. 
Since the manner of constructing the structure matrix $\mathbf{Z}_\emph{S}$ is critical, we investigate various construction methods in Section \ref{sec: ablation}. In addition, based on the observation that the geometric structure information of a typical sample can be reconstructed by those of other samples within the same subspace, we argue that $\mathbf{Z}_\emph{S}$ should also meet the SE priori. Therefore, we introduce another fully-connected layer without the bias and activation function to mimic the SE process on $\mathbf{Z}_\emph{S}$. By imposing the Frobenius norm, the structure SE formulation can be written as
\begin{equation}
\begin{aligned}
& \min_{\mathbf{C}_\emph{S}}
  \lambda_1 \left\| \mathbf{C}_\emph{S} \right\|^2_F 
+ \lambda_2 \left\| \mathbf{Z}_\emph{S} - \mathbf{Z}_\emph{S}\mathbf{C}_\emph{S} \right\|^2_F, 
\label{eq: structure_Frob}
\end{aligned}
\end{equation}
where $\mathbf{C}_\emph{S}\in\mathbb{R}^{n\times n}$ is the learned affinity graph from $\mathbf{Z}_\emph{S}$, $\left\| \mathbf{Z}_\emph{S} - \mathbf{Z}_\emph{S}\mathbf{C}_\emph{S} \right\|^2_F$ measures the structure SE loss, and $\lambda_1$ and $\lambda_2$ are two trade-off parameters. Here, we drop the constraint $diag(\mathbf{C}_\emph{S})=\mathbf{0}$, whose reason is the same as that in Eq. (\ref{eq: attribute_Frob}).

\subsection{Attention-based Fusion Module}
After obtaining $\mathbf{C}_\emph{A}$ and $\mathbf{C}_\emph{S}$, we fuse them together to construct the final affinity graph for spectral clustering. Here, we propose a novel attention-based fusion module. Specifically, we first concatenate $\mathbf{C}_\emph{A}$ and $\mathbf{C}_\emph{S}$ as $[\mathbf{C}_\emph{A}~\mathbf{C}_\emph{S}]\in\mathbb{R}^{n\times {2n}}$, and then introduce a weight matrix $\mathbf{W}\in\mathbb{R}^{{2n}\times 2}$ by learning their attentional coefficients to capture the relationship between $\mathbf{C}_\emph{A}$ and $\mathbf{C}_\emph{S}$, \textcolor{black}{where $\mathbf{W}$ is initialized with the all-ones initialization to endow the same weight to $\mathbf{C}_\emph{A}$ and $\mathbf{C}_\emph{S}$ at the beginning of training}. After that, the LeakyReLU \cite{maas2013rectifier} activation is applied to the product of $\left[\mathbf{C}_\emph{A}~\mathbf{C}_\emph{S}\right]$ and $\mathbf{W}$, where the negative input slope of LeakyReLU is set as $0.2$. We then normalize the output of the LeakyReLU unit via a softmax function and an $\ell_2$ normalization. The overall process is written as 
\begin{equation}
\begin{aligned}
\mathbf{M}=\ell_{2}\left( \rm{softmax} \left( \rm{LeakyReLU}\left(\left[\mathbf{C}_\emph{A}~\mathbf{C}_\emph{S}\right]\mathbf{W}\right)\right)\right),
\label{eq: AF_A}
\end{aligned}
\end{equation}
where $\mathbf{M}=[\mathbf{m}_{1}~ \mathbf{m}_{2}]\in\mathbb{R}^{n\times 2}$ is a weight matrix with entries being greater than $0$, and $\mathbf{m}_{1}\in\mathbb{R}^{n\times 1}$ and $\mathbf{m}_{2}\in\mathbb{R}^{n\times 1}$ are the weight vectors for measuring the importance of $\mathbf{C}_\emph{A}$ and $\mathbf{C}_\emph{S}$, respectively. Accordingly, we adaptively fuse $\mathbf{C}_\emph{A}$ and $\mathbf{C}_\emph{S}$ as
\begin{equation}
\begin{aligned}
\mathbf{C}_\emph{F}
& = \emph{f}_\emph{AF}\left( \mathbf{C}_\emph{A}, \mathbf{C}_\emph{S} \right)= \left( \mathbf{m}_{1}\mathbf{1} \right) \odot \mathbf{C}_\emph{A} + \left( \mathbf{m}_{2}\mathbf{1} \right) \odot \mathbf{C}_\emph{S},
\label{eq: AF}
\end{aligned}
\end{equation}
where $\mathbf{C}_\emph{F}\in\mathbb{R}^{n\times n}$ is the fused affinity matrix, $\mathbf{1}\in\mathbb{R}^{1\times n}$ is a vector with all entries being one, and $\odot$ denotes the Hadamard product of matrices. 
Afterward, we exploit $\mathbf{C}_\emph{F}$ to construct a similarity matrix $\mathbf{S}=f_1(\mathbf{C}_\emph{F})$ with its element
\begin{equation}
\emph{s}_{i,\emph{j}} = \left\{\begin{array}{ll}
\left(\emph{c}_{i,\emph{j}}+\emph{c}_{j,\emph{i}}\right)/\left(2*c_{sum}\right) & \text { if } i \neq j, \\
1 & \text { otherwise,}
\end{array}\right.
\end{equation}
where $\emph{c}_{i,\emph{j}}$ is the $(\emph{i}, \emph{j})$-th element of $\mathbf{C}_\emph{F}$, and $c_{sum}$ is the sum of off-diagonal entries of the current row. 
Similar to \cite{xie2016unsupervised}, we use the high-confidence elements of $\mathbf{C}_\emph{F}$ to train the attention-based fusion module by constructing an auxiliary target distribution $\mathbf{P}=f_2(\mathbf{C}_\emph{F})$ with its element
\begin{equation}
\begin{aligned}
p_{i,\emph{j}} = \left(s_{i,\emph{j}}^{2}/\sum_{i} s_{i,\emph{j}}\right)/\left(\sum_{\emph{j}^{'}} s_{i,\emph{j}^{'}}^{2}/\sum_{i} s_{i,\emph{j}^{'}}\right).
\end{aligned}
\end{equation}
We regard $\mathbf{S}$ as a special distribution and minimize the Kullback-Leibler (KL) divergence between $\mathbf{P}$ and $\mathbf{S}$, i.e., 
\begin{equation}
\begin{aligned}
KL(\mathbf{P}, \mathbf{S})\!=\!KL(f_2(\mathbf{C}_\emph{F}),f_1(\mathbf{C}_\emph{F}))\!=\!\sum_i\sum_j{p_{i,\emph{j}} log{\frac{p_{i,\emph{j}}}{s_{i,\emph{j}}}}},
\label{eq: KL}
\end{aligned}
\end{equation}
to guide network training. We can finally utilize $\mathbf{C}_F$ to conduct the spectral clustering to obtain the clustering results.

\subsection{Training Settings}
Similar to \cite{zhou2018deep,lv2021pseudo}, we first pre-train the proposed network and then fine-tune it.

\subsubsection{Pre-training Stage} 
As the network is unsupervised and challenging to infer its parameters, we first singly pre-train the feature representation module (i.e., without the dual SE-based module and the attention-based fusion module) to obtain a good initialization for the network parameters. Specifically, we only minimize the AE reconstruction loss function in Eq. (\ref{eq: AE}) with a gradient descent method (i.e., ADAM \cite{kingma2014adam}).

\subsubsection{Fine-tuning Stage} 
After pre-training, we combine Eqs. (\ref{eq: AE}), (\ref{eq: attribute_Frob}), (\ref{eq: structure_Frob}), (\ref{eq: AF}), and (\ref{eq: KL}) to form the overall loss function:
\begin{equation}
\begin{aligned}
\min_{\mathbf{C}_\emph{F}}
& \frac{1}{2} \!\left\| \mathbf{X}\!-\!\hat{\mathbf{X}} \right\|^2_F 
\!\!+\!\left\| \mathbf{C}_\emph{A} \right\|^2_F 
\!+\!\gamma\!\left\| \mathbf{Z}_\emph{A}\!-\!\mathbf{Z}_\emph{A}\mathbf{C}_\emph{A} \right\|^2_F\!+\!\lambda_1\!\left\| \mathbf{C}_\emph{S} \right\|^2_F \\
& 
\!+\!\lambda_2 \left\| \mathbf{Z}_\emph{S} - \mathbf{Z}_\emph{S}\mathbf{C}_\emph{S} \right\|^2_F 
\textcolor{black}{ + \beta KL(f_2(\mathbf{C}_\emph{F}),f_1(\mathbf{C}_\emph{F}))}\\
\rm{s.t.} & \ \mathbf{Z}_{\emph{A}} \!= \emph{f}_\emph{AE}\left( \mathbf{X}\right), \mathbf{Z}_{\emph{S}} \!= \frac{\mathbf{C}_\emph{A}+\mathbf{C}_\emph{A}^\mathsf{T}}{2}, \mathbf{C}_\emph{F} \!= \emph{f}_\emph{AF}\left( \mathbf{C}_\emph{A}, \mathbf{C}_\emph{S} \right),
\label{eq: Our}
\end{aligned}
\end{equation}
and fine-tune the complete network by optimizing Eq. (\ref{eq: Our}). 

\section{Experiments}
\label{sec: eprm}
We quantitatively and qualitatively compared the proposed method with eighteen state-of-the-art methods on five commonly used benchmark datasets. In addition, we performed extensive ablation studies of the designed modules and examined the influence of the hyperparameters. 

\subsection{Datasets}
We conducted the experiments on UMIST\footnote{http://eprints.lincoln.ac.uk/id/eprint/16081/}, ORL\footnote{http://www.cl.cam.ac.uk/research/dtg/attarchive/facedatabase.html}, MNIST\footnote{http://yann.lecun.com/exdb/mnist}, COIL20\footnote{http://www.cs.columbia.edu/CAVE/software/softlib/coil-20.php}, and COIL40\footnote{https://github.com/sckangz/L2SP}, and the detailed descriptions of these datasets are as follows.

\begin{itemize}
\item  \textbf{UMIST} \cite{graham1998characterising}. The University of Manchester Institute of science and technology (UMIST) database contains 564 images of 20 individuals, in which each individual presents a range of poses from profile to frontal views. In this paper, we used 480 images of 20 people and down-sampled each image to $32 \times 32$.
\item  \textbf{ORL} \cite{samaria1994parameterisation}. The Olivetti research ltd database contains 40 individuals with 400 images, in which each image is captured at different scenes (e.g., the lighting condition). 
\item  \textbf{MNIST} \cite{lecun1998gradient}. The modified national institute of standards and technology database contains ten digits with 60000 training examples and 10000 test examples. In this paper, we used the first 100 images of each digit to conduct experiments.
\item	\textbf{COIL20, COIL40} \cite{zhou2018deep}. The Columbia object image library dataset contains COIL20 and COIL40, in which COIL20 has 1440 images of 20 objects, and COIL40 has 2880 images of 40 objects. Here, each object has 72 views, and all images are down-sampled to $32 \times 32$.
\end{itemize}

\begin{table}[]
\caption{The details of the network architecture on five datasets.} 
\label{tab: NN setting}
\centering
\setlength{\tabcolsep}{1.28mm}{
\begin{tabular}{c|c|c|c}
\hline\hline
Dataset & Layers      & Encoder                           & Decoder                                 \\
\hline\hline
\multirow{2}{*}{UMIST}   & Kernel size & (5$\times$5, 3$\times$3, 3$\times$3) & (3$\times$3, 3$\times$3, 5$\times$5) \\
        & Channel     & (15, 10, 5)                          & (5, 10, 15)                          \\
\hline
\multirow{2}{*}{ORL}     & Kernel size & (3$\times$3, 3$\times$3, 3$\times$3) & (3$\times$3, 3$\times$3, 3$\times$3) \\
        & Channel     & (3, 3, 5)                            & (5, 3, 3)                            \\
\hline
\multirow{2}{*}{MNIST}   & Kernel size & (5$\times$5, 3$\times$3, 3$\times$3) & (5$\times$5, 3$\times$3, 3$\times$3) \\
        & Channel     & (10, 20, 30)                         & (30, 20, 10)                         \\
\hline
\multirow{2}{*}{COIL20}  & Kernel size & 3$\times$3                           & 3$\times$3                           \\
        & Channel     & 15                                   & 15                                   \\
\hline
\multirow{2}{*}{COIL40}  & Kernel size & 3$\times$3                           & 3$\times$3                           \\
        & Channel     & 20                                   & 20                                   \\
\hline\hline
\end{tabular}
}
\end{table}

\begin{table*}[]
\caption{Clustering results on benchmark datasets. The best and second-best results are \textbf{bolded} and \underline{underlined}, respectively.}
\centering
\setlength{\tabcolsep}{1.28mm}{
\footnotesize
\label{tab: final_result}
\begin{tabular}{c|ccc|ccc|ccc|ccc|ccc}
\hline\hline
Datasets           & \multicolumn{3}{c|}{UMIST}               & \multicolumn{3}{c|}{ORL}              & \multicolumn{3}{c|}{MNIST}              & \multicolumn{3}{c|}{COIL20}             & \multicolumn{3}{c}{COIL40}         \\
\hline
Methods\textbackslash{}Metrics & ACC       & NMI       & PUR       & ACC       & NMI       & PUR       & ACC       & NMI       & PUR       & ACC       & NMI       & PUR       & ACC       & NMI       & PUR   \\
\hline\hline
\rowcolor{gray!40}SSC \cite{elhamifar2009sparse}
& 0.6904             & 0.7489             & 0.6554             & 0.7425             & 0.8459             & 0.7875             & 0.4530             & 0.4709             & 0.4940            & 0.8631             & 0.8892             & 0.8747             & 0.7191             & 0.8212             & 0.7716 \\
EDSC \cite{ji2014efficient}
& 0.6937             & 0.7522             & 0.6683             & 0.7038             & 0.7799             & 0.7138             & 0.5650             & 0.5752             & 0.6120             & 0.8371             & 0.8828             & 0.8585             & 0.6870             & 0.8139             & 0.7469\\
\rowcolor{gray!40}KSSC \cite{patel2014kernel}
& 0.6531             & 0.7377             & 0.6256             & 0.7143             & 0.8070             & 0.7513             & 0.5220             & 0.5623             & 0.5810             & 0.7087             & 0.8243             & 0.7497             & 0.6549             & 0.7888             & 0.7284    \\
SSC-OMP \cite{you2016scalable}
& 0.6438             & 0.7068             & 0.6171             & 0.7100             & 0.7952             & 0.7463             & 0.3400             & 0.3272             & 0.3560            & 0.6410             & 0.7412             & 0.6667             & 0.4431             & 0.6545             & 0.5250    \\
\rowcolor{gray!40}ENSC \cite{you2016oracle}                   
& 0.6931             & 0.7569             & 0.6628             & 0.7525             & 0.8540             & 0.7950             & 0.4983             & 0.5495             & 0.5483            & 0.8760             & 0.8952             & 0.8892             & 0.7426             & 0.8380             & 0.7924   \\
LRR \cite{liu2012robust}                                      
& 0.6979             & 0.7630             & 0.6670             & 0.8100             & 0.8603             & 0.8225             & 0.5386             & 0.5632             & 0.5684          & 0.8118             & 0.8747             & 0.8361             & 0.6493             & 0.7828             & 0.7109       \\
\rowcolor{gray!40}LRSC \cite{vidal2014low}                  
& 0.6729             & 0.7498             & 0.6562             & 0.7200             & 0.8156             & 0.7542             & 0.5140             & 0.5576             & 0.5550                       & 0.7416             & 0.8452             & 0.7937             & 0.6327             & 0.7737             & 0.6981    \\
FLSR \cite{ma2020towards}                                     
& 0.6000             & 0.7082             & 0.6188             & 0.7775             & 0.8661             & 0.7900             & 0.6210             & 0.5231             & 0.6210                           & 0.7104             & 0.7847             & 0.7507             & 0.5920             & 0.7554             & 0.6227 \\
\rowcolor{gray!40}FTRR \cite{ma2020towards} 
& 0.7667             & 0.8509             & 0.7917             & 0.8600             & 0.9151             & 0.8725             & 0.7070             & 0.6672             & 0.7070                           & 0.9035             & 0.9305             & 0.9104             & 0.7858             & 0.8801             & 0.8156       \\
\hline
AE \cite{hinton2006reducing} + SSC                                    
& 0.7042           & 0.7515             & 0.6785             & 0.7563             & 0.8555             & 0.7950             & 0.4840             & 0.5337             & 0.5290                          & 0.8711             & 0.8990             & 0.8901             & 0.7391             & 0.8318             & 0.7840            \\
\rowcolor{gray!40}DEC \cite{xie2016unsupervised}                                      
& 0.5521             & 0.7125             & 0.5917             & 0.5175             & 0.7449             & 0.5400             & 0.6120             & 0.5743             & 0.6320           & 0.7215             & 0.8007             & 0.6931             & 0.4872             & 0.7417             & 0.4163         \\
DKM \cite{fard2020deep}                    
& 0.5106             & 0.7249             & 0.5685             & 0.4682             & 0.7332             & 0.4752             & 0.5332             & 0.5002             & 0.5647           & 0.6651             & 0.7971             & 0.6964             & 0.5812             & 0.7840             & 0.6367          \\
\rowcolor{gray!40}DCCM \cite{wu2019deep}                                      
& 0.5458             & 0.7440             & 0.5854             & 0.6250             & 0.7906             & 0.5975             & 0.4020             & 0.3468             & 0.4370             & 0.8021             & 0.8639             & 0.7889             & 0.7691             & 0.8890             & 0.7663         \\
DEPICT \cite{ghasedi2017deep}                 
& 0.4521             & 0.6329             & 0.4167             & 0.2800             & 0.5764             & 0.1450             & 0.4240             & 0.4236             & 0.3560             & 0.8618             & 0.9266             & 0.8319             & 0.8073             & \underline{0.9291}             & 0.8191 \\
\rowcolor{gray!40}DSC-Net-L1  \cite{ji2017deep}             
& 0.7242             & 0.7556             & 0.7204             & 0.8550             & 0.9023             & 0.8585             & 0.7280             & 0.7217             & 0.7890                         & 0.9314             & 0.9353             & 0.9306             & 0.8003             & 0.8852             & 0.8646     \\
DSC-Net-L2  \cite{ji2017deep}                               
& 0.7312             & 0.7662             & 0.7276             & 0.8600             & 0.9034             & 0.8625             & 0.7500             & 0.7319             & 0.7991             & 0.9368             & 0.9408             & 0.9397             & 0.8075             & 0.8941             & 0.8740          \\
\rowcolor{gray!40}DASC \cite{zhou2018deep}
& 0.7688             & 0.8042             & 0.7688             & \underline{0.8825} & 0.9315             & \underline{0.8925} & 0.8040             & \textbf{0.7800}             & 0.8370           & 0.9639             & 0.9686             & 0.9632             & 0.8354             & 0.9196             & \textbf{0.8972}  \\
PSSC \cite{lv2021pseudo}
& \underline{0.7917} & \underline{0.8670} & \underline{0.8146} & 0.8675  & \underline{0.9349} & {0.8925}   & \underline{0.8430} & \underline{0.7676}          & \underline{0.8430}   & \underline{0.9722} & \underline{0.9779} & \underline{0.9722} & \underline{0.8358} & {0.9258} & 0.8642     \\
\hline
\rowcolor{gray!40}
\textbf{Our} & \textbf{0.8354} & \textbf{0.8902} & \textbf{0.8500}    & \textbf{0.9075}    & \textbf{0.9431}    & \textbf{0.9175}    & \textbf{0.8460}    & {0.7609} & \textbf{0.8460} &  \textbf{0.9840}    & \textbf{0.9829}    & \textbf{0.9840}    & \textbf{0.8719}    & \textbf{0.9343}    & \underline{0.8899}  \\
\hline\hline
\end{tabular}
}
\end{table*}

\subsection{Compared Methods}
We compared the proposed method with eighteen state-of-the-art methods, and the detailed descriptions of these methods are as follows.
\begin{itemize}
	\item	\textbf{SSC} \cite{elhamifar2009sparse,elhamifar2013sparse} imposes a sparse regularization to seek a subspace-preserving data affinity.
	\item	\textbf{EDSC} \cite{ji2014efficient} yields dense connections within each cluster.
	\item	\textbf{KSSC} \cite{patel2014kernel} extends SSC to nonlinear manifold by employing a kernel function.
	\item	\textbf{SSC-OMP} \cite{you2016scalable} introduces the orthogonal matching pursuit to SSC.
    \item	\textbf{ENSC} \cite{you2016oracle} balances the subspace-preserving (sparse solutions) and the connectivity (dense solutions) by using an elastic net.
	\item	\textbf{LRR} \cite{liu2012robust} seeks the lowest rank representation by exploiting a low rank regularization.
	\item	\textbf{LRSC} \cite{vidal2014low} introduces the low rank regularization to the subspace clustering to solve the corrupted data (e.g., noise or errors).
	\item \textbf{FLSR}, \textbf{FTRR} \cite{ma2020towards} inject graph similarity into data features based on least square regression \cite{lu2012robust} and thresholding ridge regression \cite{peng2015robust}, respectively.
	\item	\textbf{AE+SSC} conducts SSC on a pre-trained auto-encoder (AE) \cite{hinton2006reducing} feature.
	\item	\textbf{DEC} \cite{xie2016unsupervised} partitions data points in a jointly optimized feature space.
	\item	\textbf{DKM} \cite{fard2020deep} achieves clustering by jointly learning a deep representation and performing k-means \cite{hartigan1979algorithm}.
	\item	\textbf{DCCM} \cite{wu2019deep} learns feature representation by mining comprehensive structures for deep unsupervised clustering.
	\item	\textbf{DEPICT} \cite{ghasedi2017deep} conducts a nonlinear projection and predicts cluster assignments with a regularized relative entropy loss function.
	\item  \textbf{DSC-Net-L1}, \textbf{DSC-Net-L2} \cite{ji2017deep} are two deep subspace clustering networks with sparse \cite{elhamifar2009sparse} and Tikhonov \cite{lu2012robust} regularization, respectively.
	\item	\textbf{DASC} \cite{zhou2018deep} supervises sample representation learning and subspace clustering in an adversarial learning manner.
	\item	\textbf{PSSC} \cite{lv2021pseudo} conducts the subspace clustering by integrating locality preserving, self-expression, and self-supervision.
\end{itemize}

\subsection{Evaluation Metrics}
We used three commonly used metrics (i.e., accuracy (ACC), normalized mutual information (NMI), and purity (PUR)) to quantitatively evaluate clustering performance. The larger the values of these metrics, the better the clustering performance. Specifically, ACC is calculated by
\begin{equation}
\begin{aligned}
& \text{ACC} = \frac{1}{n}\sum_{i=1}^{n} \mathbf{1} \left\lbrace y_i = m(c_i) \right\rbrace ,
\label{eq:ACC}
\end{aligned}
\end{equation}
where $y_i$ denotes the ground truth, $c_i$ denotes the predicted assignment, $m(\cdot)$ enumerates the mapping between $y_i$ and $c_i$ with the Kuhn-Munkres algorithm \cite{munkres1957algorithms}, and $\mathbf{1}\left\lbrace \cdot \right\rbrace$ denotes the indicator function with $1$ or $0$. NMI is calculated by
\begin{equation}
\begin{aligned}
& \text{NMI}=\frac{\sum_{y\in Y, \hat{y} \in \hat{Y}} p(y,\hat{y})\log\left( \frac{p(y,\hat{y})}{p(y)p(\hat{y})}\right)}{\max\left( H(Y), H(\hat{Y}) \right) },
\label{eq:NMI}
\end{aligned}
\end{equation}
where $y$ and $\hat{y}$ denote two clusters labels, $Y$ and $\hat{Y}$ are their clusters sets, $p(\cdot)$ is a marginal probability mass function, $p(y,\hat{y})$ denotes a joint probability mass function of $Y$ and $\hat{Y}$, and $H(\cdot)$ represents an entropy function \cite{shannon1948mathematical}. PUR is calculated by
\begin{equation}
\begin{aligned}
& \text{PUR}=\frac{1}{n}\sum_{i=1}^{k}\max_{j}|c_i\cap t_j|,
\label{eq:PUR}
\end{aligned}
\end{equation}
where $t_j$ denotes the segmentation which has the max count for $c_i$, $\max_{j}|c_i\cap t_j|$ counts the number of data points from the most common class in $c_i$, $k$ denotes the number of clusters. 

\subsection{Implementation Details}
Table \ref{tab: NN setting} lists the detailed network settings of our method on each dataset. In addition, we set the kernel stride of both horizontal and vertical directions to be $2$ and used a rectified linear unit (ReLU) \cite{krizhevsky2012imagenet} activation function. We implemented the proposed method with TensorFlow 2.1.0 and initialized the neural network parameters by the `he\_normal' initializer \cite{he2015delving}. We followed the  decoupling design in \cite{peng2021maximum} to construct our overall framework. Besides, we set $\gamma$ as the same as the optimal parameter of \cite{ji2017deep}. After training, we performed spectral clustering on the fused affinity graph. For a fair comparison, we used the normalized cut algorithm \cite{shi2000normalized} as in \cite{ji2017deep,zhou2018deep,ma2020towards,lv2021pseudo}. 

\subsection{Clustering Results}
Table \ref{tab: final_result} lists the quantitative results of all methods over five benchmark datasets, where we have the following conclusions. 
\begin{itemize} 
  \item The deep SE-based methods achieve better clustering performance than the traditional SE-based methods, indicating that the nonlinear projection is helpful to extract a more clustering-friendly representation to deal with complex real-world datasets. For example, on  MNIST, our method improves the ACC, NMI, and PUR values of the best traditional SE-based approach FTRR by 13.90\%, 9.37\%, and 13.90\%, respectively.
  \item Structure information contributes to achieving better clustering performance. Specifically, FTRR, which exploits structure information to enhance the feature representation, produces the best performance among those traditional SE-based methods, e.g., on COIL40, FTRR obtains 4.32\% ACC improvement over the second-best traditional SE-based method ENSC. In addition, our method performs best among all deep SE-based methods, e.g., on COIL40, our AASSC-Net improves the ACC value of the second-best method PSSC by 3.61\%. 
  \item Simultaneously considering the attribute and structure information is beneficial to clustering, which is validated by comparing DSC-Net-L2 and our method. 
  \item Our AASSC-Net almost always obtains the best clustering performance among all comparisons. For example, on UMIST, our approach achieves 4.37\%, 2.32\%, and 3.54\% improvements in terms of ACC, NMI, and PUR, respectively, compared with the second-best method PSSC. Such impressive performance is credited to the simultaneous consideration of the attribute and structure information via the attention-based fusion module.
\end{itemize}

\begin{figure*}[!ht]
	\centering
	\subfigure[$\mathbf{C}_\emph{A}$]{
		\includegraphics [width=0.60\columnwidth]{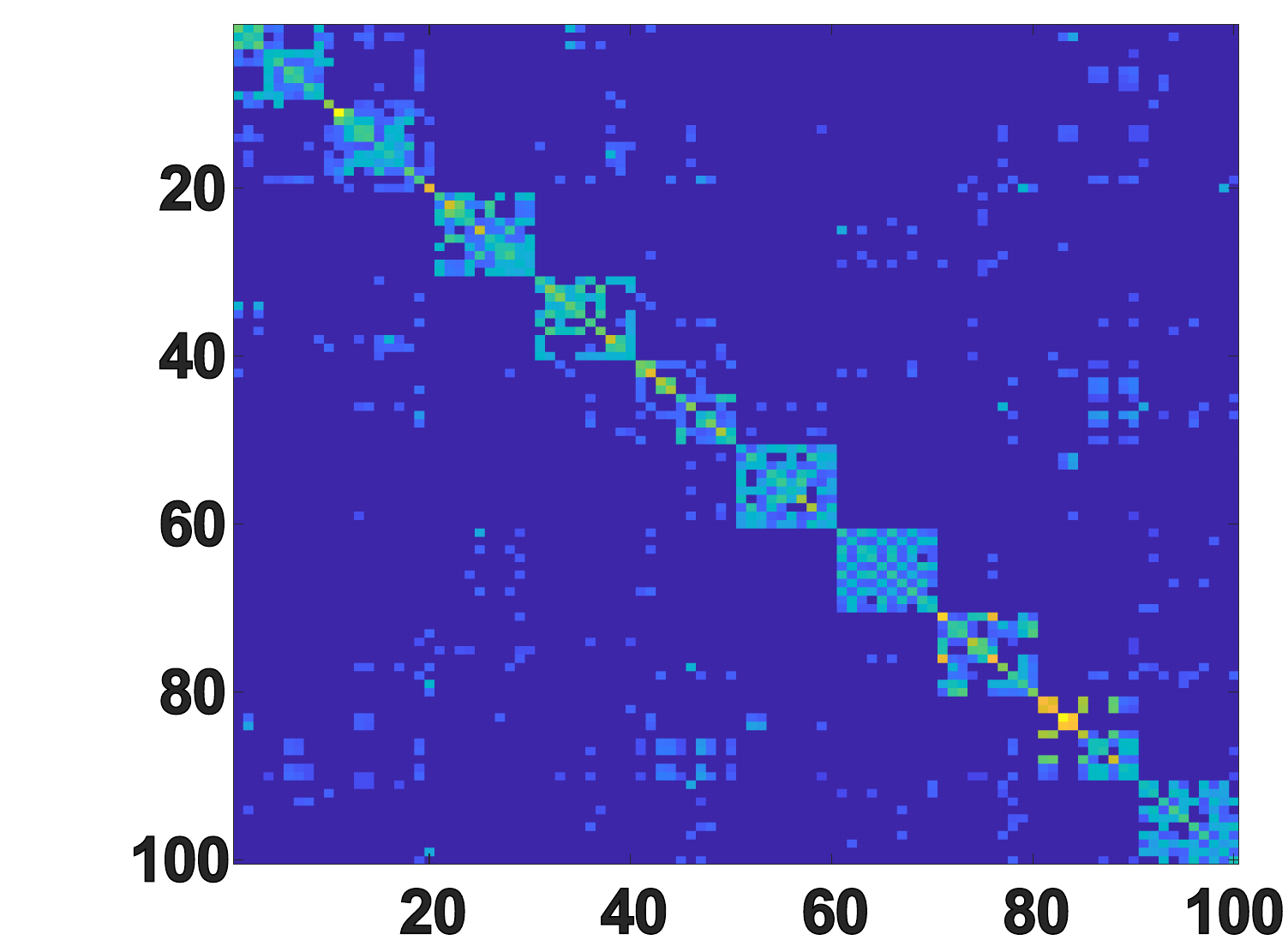}
	}
	\subfigure[$\mathbf{C}_\emph{S}$]{
		\includegraphics [width=0.60\columnwidth]{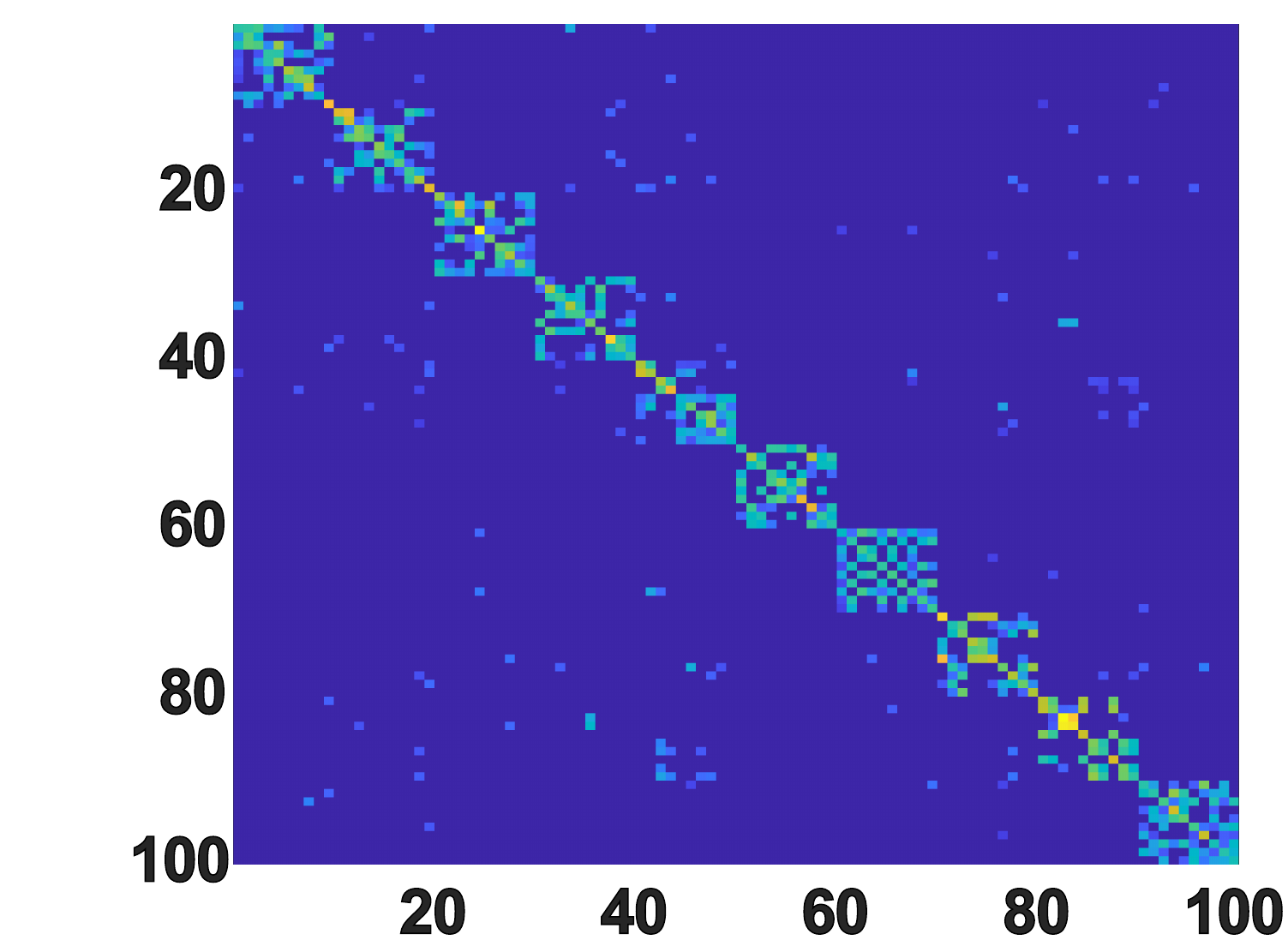}
	}
	\subfigure[$\mathbf{C}_\emph{F}$]{
		\includegraphics [width=0.60\columnwidth]{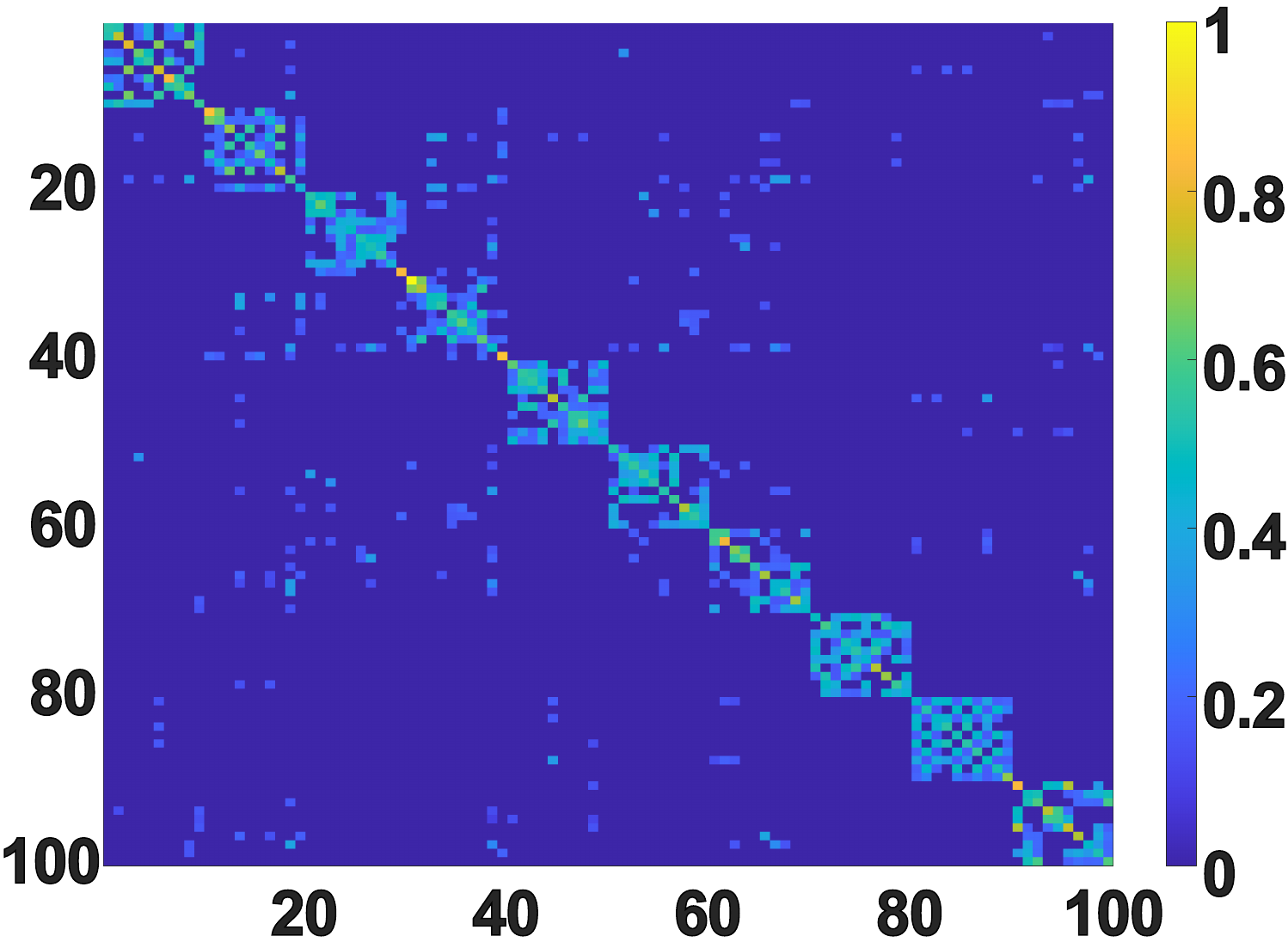}
	}
	\caption{Visual comparison of the affinity matrices $\mathbf{C}_\emph{A}$ and $\mathbf{C}_\emph{S}$ respectively learned from attribute and structure information on the synthetic dataset, as well the fused one $\mathbf{C}_\emph{F}$.}
	\label{fig: afm}
\end{figure*}
\begin{figure*}[!ht]
	\centering
	\subfigure[$\mathbf{C}_\emph{A}$ (ACC: 0.88)]{
		\includegraphics [width=0.56\columnwidth]{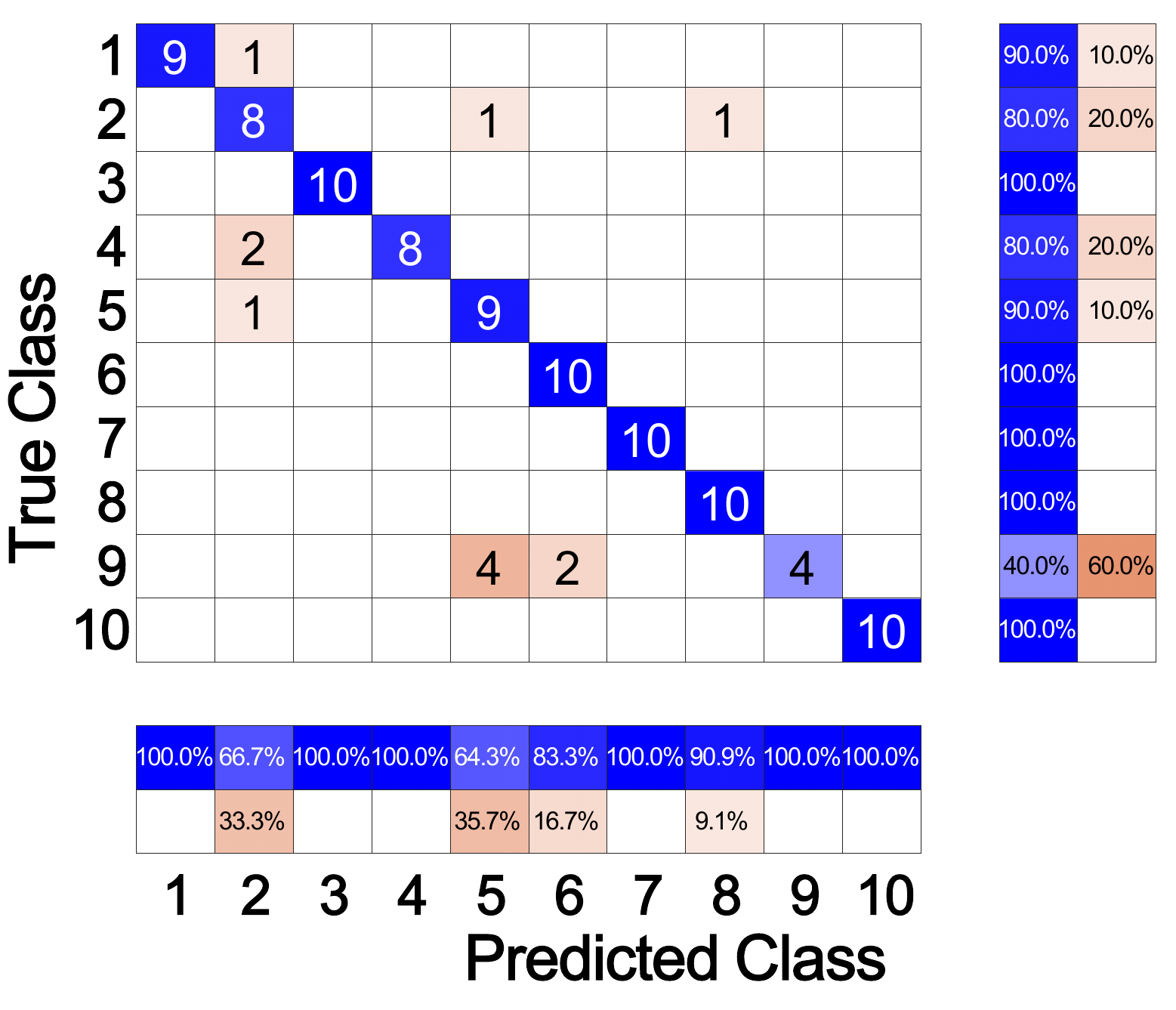}
	}
	\subfigure[$\mathbf{C}_\emph{S}$ (ACC: 0.92)]{
		\includegraphics [width=0.56\columnwidth]{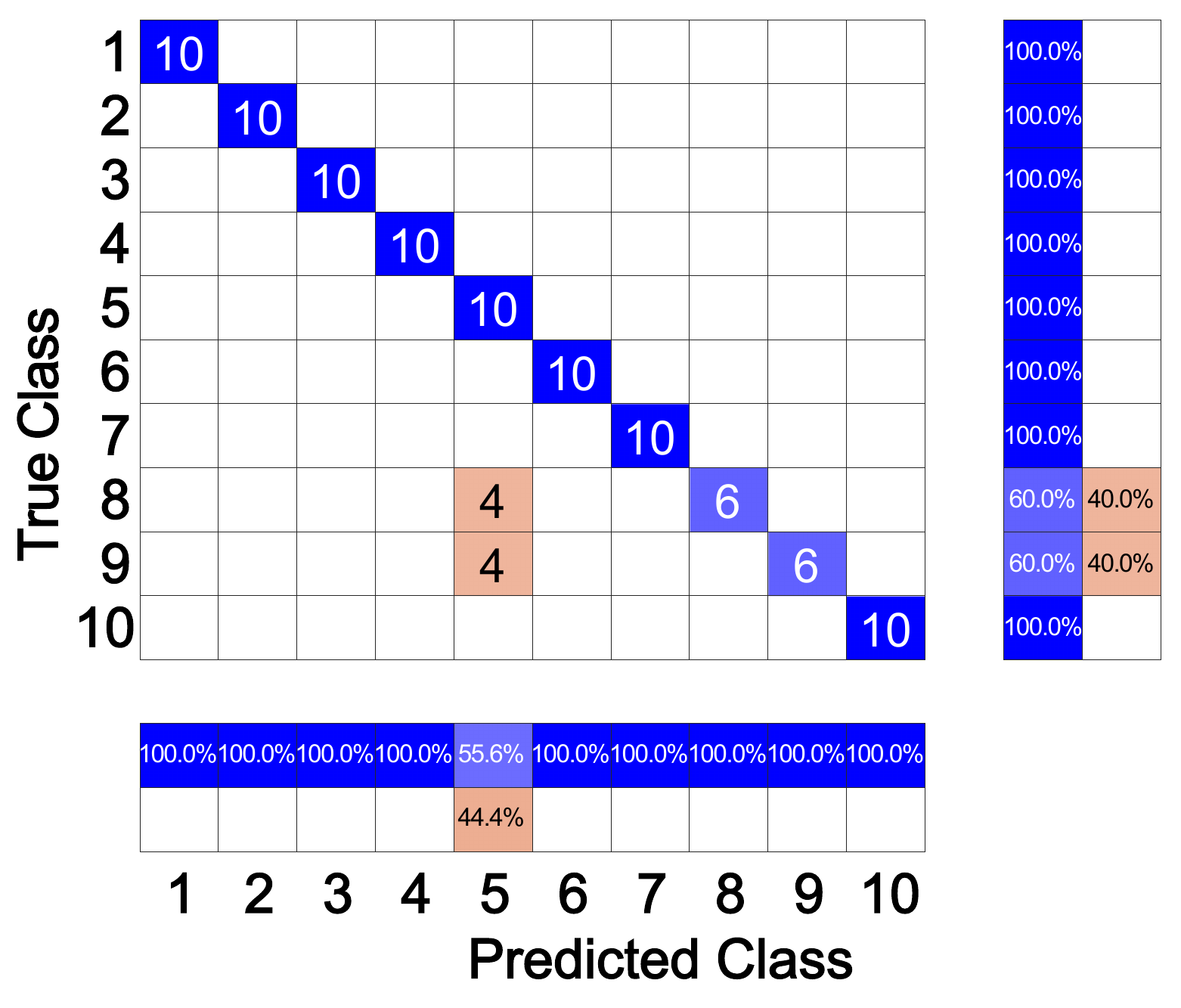}
	}
	\subfigure[$\mathbf{C}_\emph{F}$ (ACC: 0.97)]{
		\includegraphics [width=0.56\columnwidth]{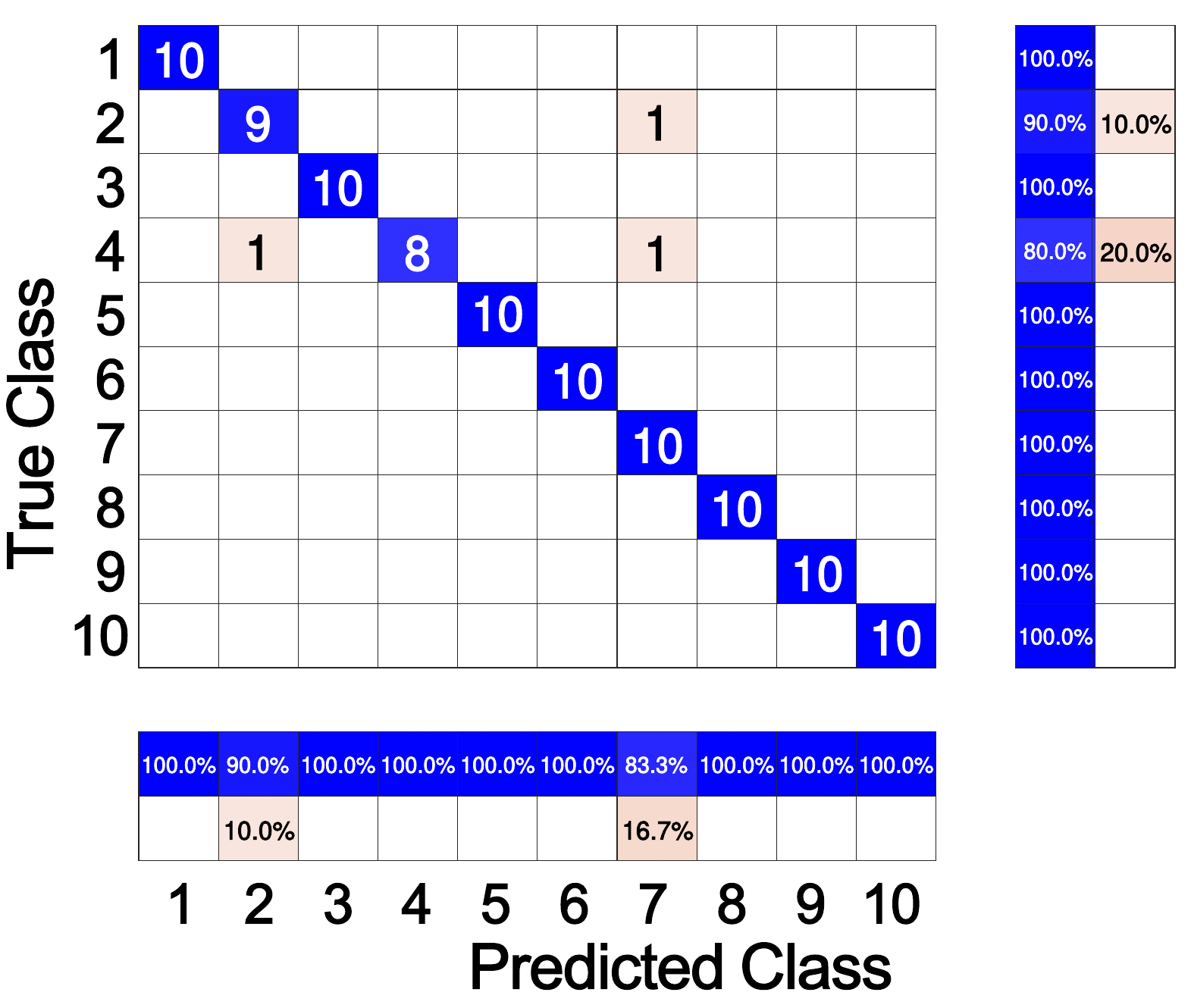}
	}
	\caption{Quantitative comparison of $\mathbf{C}_\emph{A}$, $\mathbf{C}_\emph{S}$ and $\mathbf{C}_\emph{F}$ on the synthetic dataset, where the below column summaries the precision, the right row summaries the recall.}
	\label{fig: cfm}
\end{figure*}

\begin{table}[]
\caption{Ablation studies, where the \textbf{bold} highlights the best results.}
\centering
\label{tab: AS}
\resizebox{0.42\textwidth}{!}{
\begin{tabular}{c|l|ccc}
\hline\hline
                        & Method & ACC    & NMI    & PUR    \\ 
\hline
\multirow{4}{*}{UMIST}  & (I)  & 0.8104 & 0.8730 & 0.8333 \\
                        & (II)  & 0.5708 & 0.7297 & 0.6000 \\
                        & (III)  & 0.8104 & 0.8756 & 0.8333 \\
                        & Ours       & \textbf{0.8354}    & \textbf{0.8902}    & \textbf{0.8500}       \\ 
\hline
\multirow{4}{*}{ORL}    & (I)        & 0.8650 & 0.9204 & 0.8700 \\
                        & (II)        & 0.8525 & 0.9217 & 0.8650 \\
                        & (III)        & 0.9000 & 0.9343 & 0.9000 \\
                        & Ours       &\textbf{0.9075}    & \textbf{0.9431}    & \textbf{0.9175}       \\ 
\hline
\multirow{4}{*}{MNIST}  & (I)        & 0.7890 & 0.7371 & 0.7890 \\
                        & (II)        & 0.7370 & 0.7211 & 0.7840 \\
                        & (III)        & 0.7900 & 0.7335 & 0.7900 \\
                        & Ours       & \textbf{0.8460}    & \textbf{0.7609} & \textbf{0.8460}       \\ 
\hline
\multirow{4}{*}{COIL20} & (I)        & 0.9569 & 0.9584 & 0.9569 \\
                        & (II)        & 0.9535 & 0.9589 & 0.9535 \\
                        & (III)        & 0.9806 & 0.9803 & 0.9806 \\
                        & Ours       & \textbf{0.9840}    & \textbf{0.9829}    & \textbf{0.9840} \\
\hline
\multirow{4}{*}{COIL40} & (I)        & 0.8250 & 0.9223 & 0.8465 \\
                        & (II)        & \textbf{0.8774} & \textbf{0.9382} & 0.8868 \\
                        & (III)        & 0.8608 & 0.9306 & 0.8767 \\
                        & Ours       &  0.8719    & 0.9343    & \textbf{0.8899}       \\
\hline\hline
\end{tabular}
}
\end{table}

\subsection{Ablation Study}\label{sec: ablation}

\subsubsection{Adaptive Graph Fusion} \label{subsec:ablation_graph_fusion}
To deeply understand the proposed method, we conducted comprehensive ablation studies on five benchmark datasets. 
\textcolor{black}{The experimental results are listed in Table \ref{tab: AS}, where the first row \textbf{(I)} denotes a variant of our model that only considers the module with $\mathbf{C}_\emph{A}$ to conduct spectral clustering. The second row \textbf{(II)} denotes a variant of our model that exploits the dual self-expressiveness (SE)-based module to obtain $\mathbf{C}_\emph{S}$ to conduct spectral clustering. The third row \textbf{(III)} denotes a variant of our model that simply averages $\mathbf{C}_\emph{A}$ and $\mathbf{C}_\emph{S}$, (i.e., $0.5 \ \mathbf{C}_\emph{A} + 0.5 \ \mathbf{C}_\emph{S}$), instead of using the attention-based fusion module, to conduct spectral clustering. The fourth row is our full model, i.e., \textbf{Ours}.} From Table \ref{tab: AS}, we have the following observations. 

\begin{itemize}
    \item \textcolor{black}{By comparing the results of each dataset in \textbf{(I)} and \textbf{(II)}, we can observe that $\mathbf{C}_\emph{A}$ and $\mathbf{C}_\emph{S}$ have their own advantages on clustering performance for different datasets. For instance, on COIL40, \textbf{(II)} improves 5.24\% over \textbf{(I)} on ACC, 1.59\% on NMI, and 4.03\% on PUR. However, on UMIST, \textbf{(II)} is 23.96\% worse than \textbf{(I)} on ACC.} 
    \item \textcolor{black}{The comparisons of \textbf{Ours} with \textbf{(I)} and \textbf{(II)}} of each dataset demonstrate that simultaneously considering the attribute and structure information could promote clustering performance. \textcolor{black}{For example, on MNIST, \textbf{Ours} obtains 10.90\% performance improvement over \textbf{(II)} on ACC, 3.98\% on NMI, and 6.20\% on PUR.}
    \item The advantage of the attention-based graph fusion strategy could be validated by comparing \textcolor{black}{the results of \textbf{(III)} and \textbf{Ours} of each dataset,} where it can be seen that the adaptive graph learning strategy produces a 1\% to 6\% performance improvement.
\end{itemize}

\begin{figure*}[htb]
	\centering
	\subfigure[Ground truth (1.0000)]{
		\includegraphics [width=0.47\columnwidth]{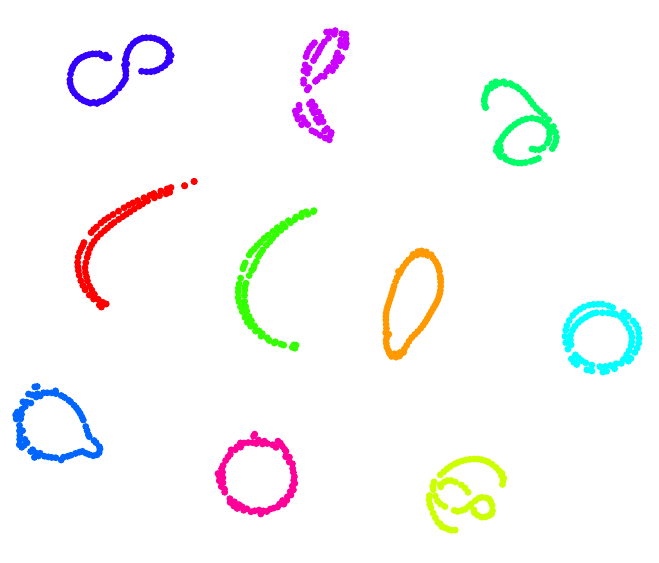}
	}
	\subfigure[DSC-Net-L2 (0.9222)]{
		\includegraphics [width=0.47\columnwidth]{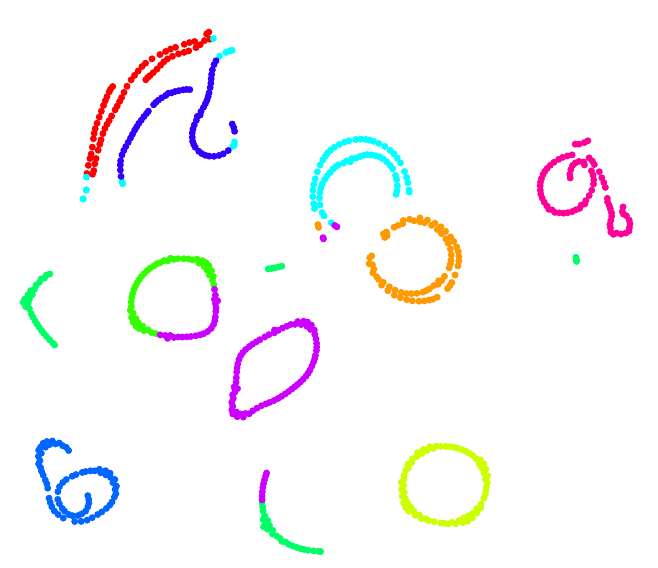}
	}
	\subfigure[PSSC (0.9514)]{
		\includegraphics [width=0.47\columnwidth]{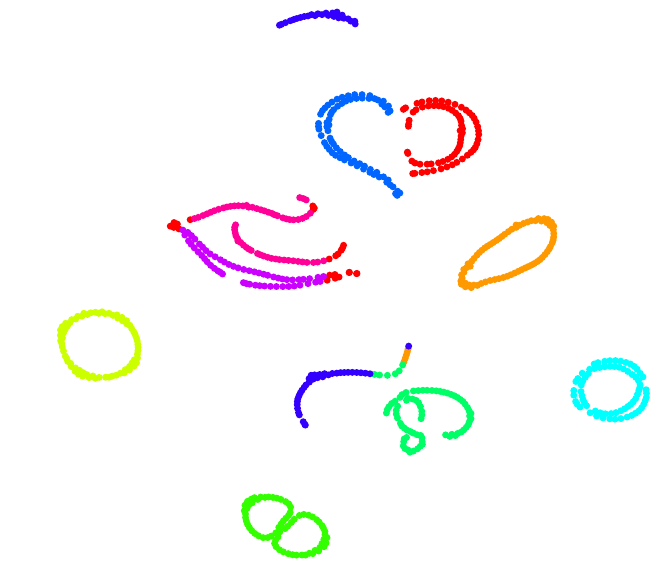}
	}
	\subfigure[{Our (\textbf{0.9903})}]{
		\includegraphics [width=0.47\columnwidth]{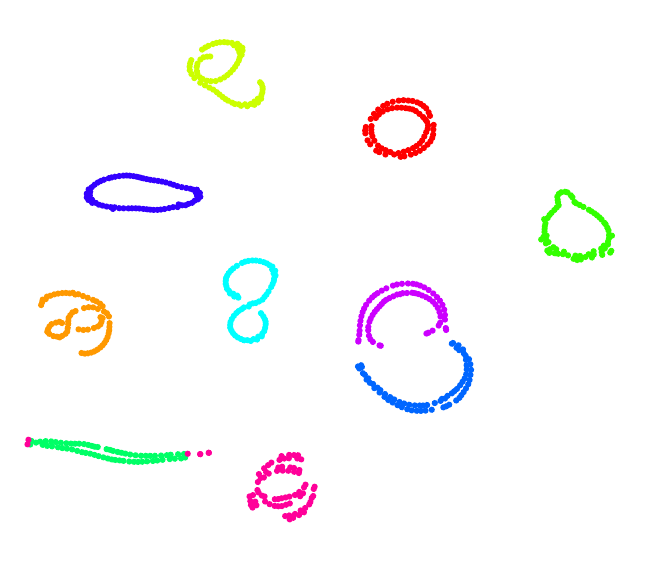}
	}
	\caption{Visualization of the learned representations by (a) ground truth, (b) DSC-Net-L2 \cite{ji2017deep}, (c) PSSC \cite{lv2021pseudo}, and (d) Our method, where the digit in the bracket refers to the ACC value. Different colors represent different clusters.}
	\label{fig:tsne}
\end{figure*}

\begin{figure*}[]
	\centering
	\subfigure[UMIST]{
	\includegraphics [width=3.10cm]{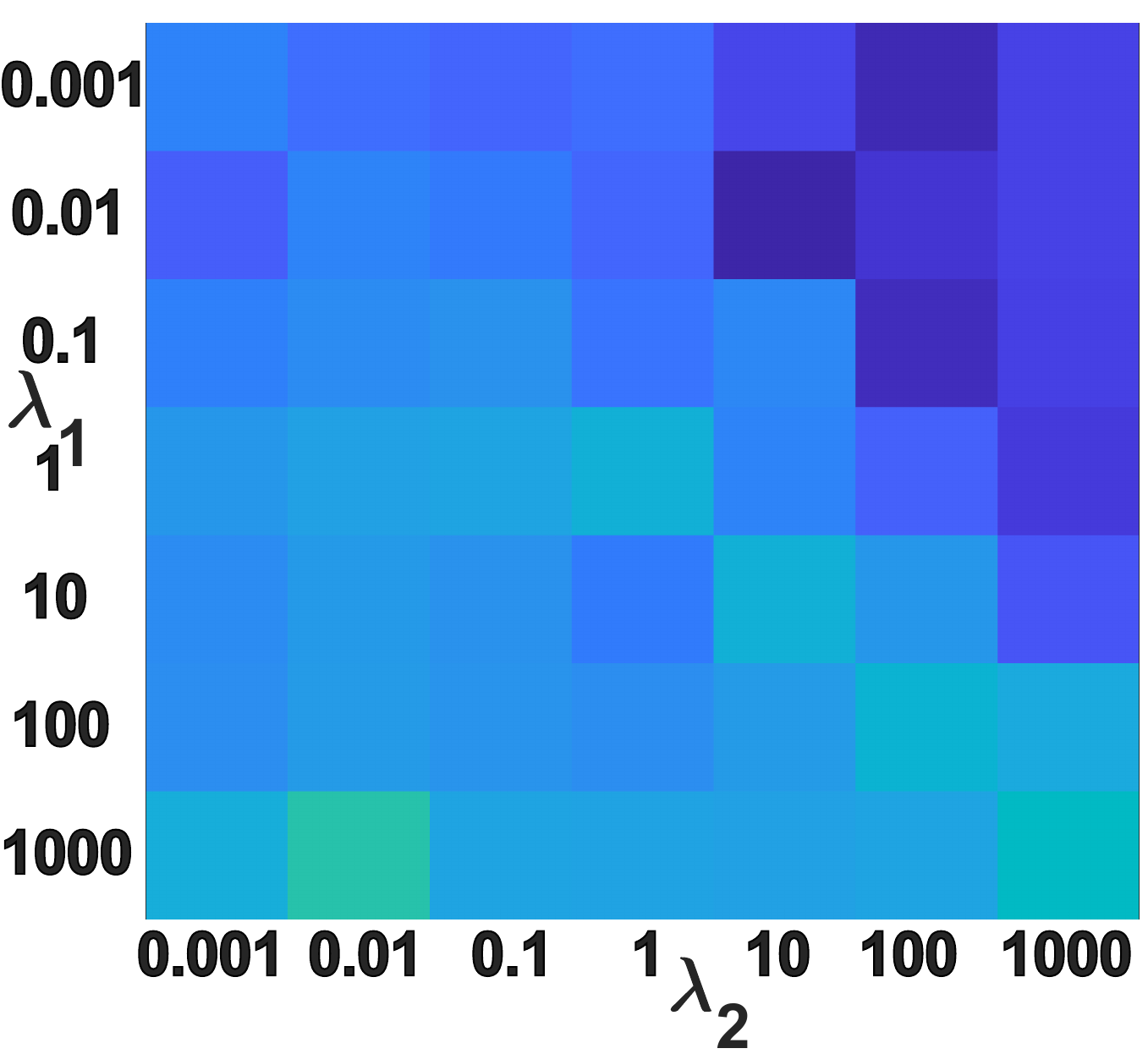}
	}
	\subfigure[ORL]{
	\includegraphics [width=3.10cm]{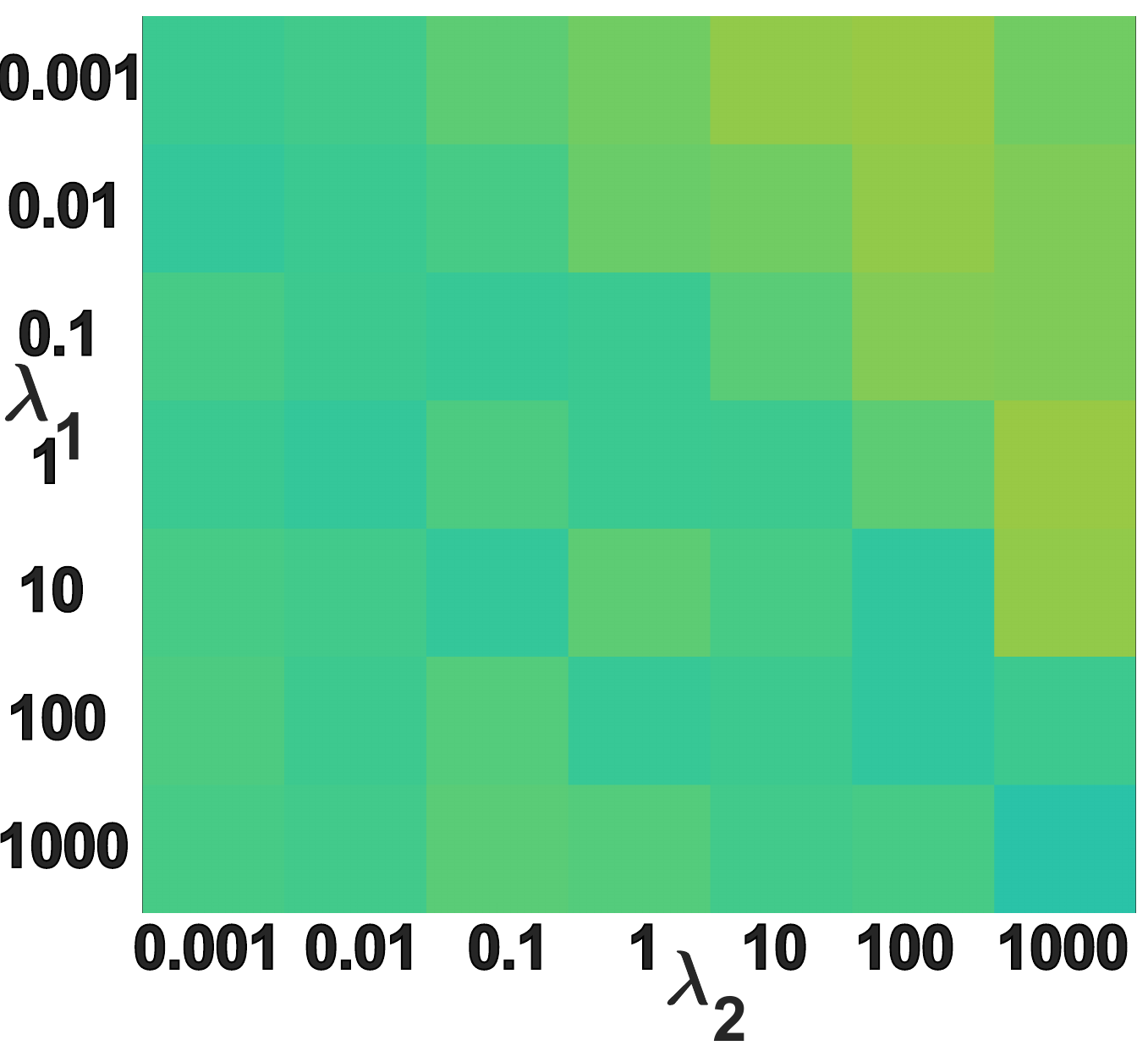}
	}
	\subfigure[MNIST]{
	\includegraphics [width=3.10cm]{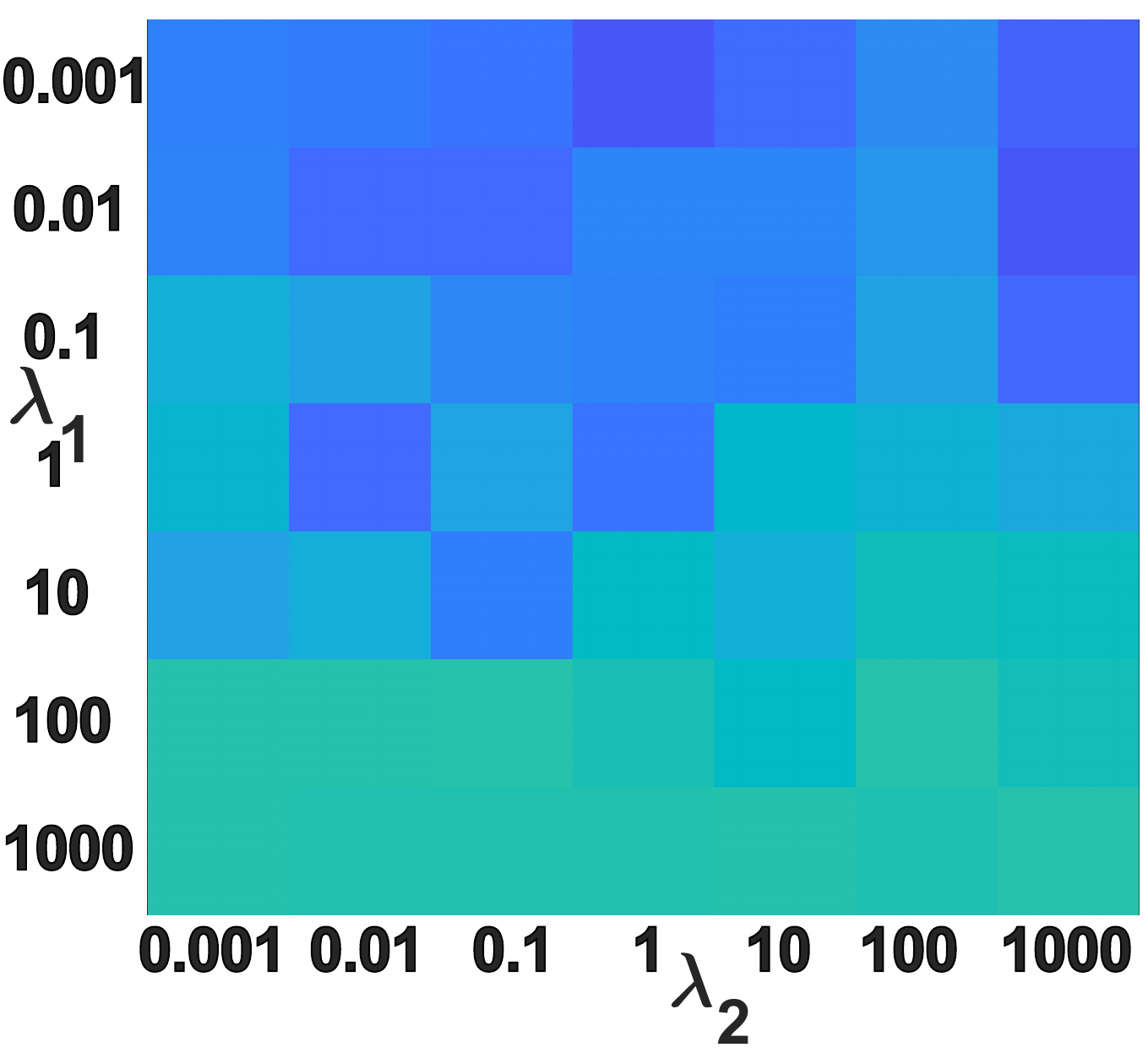}
	}
	\subfigure[COIL20]{
	\includegraphics [width=3.10cm]{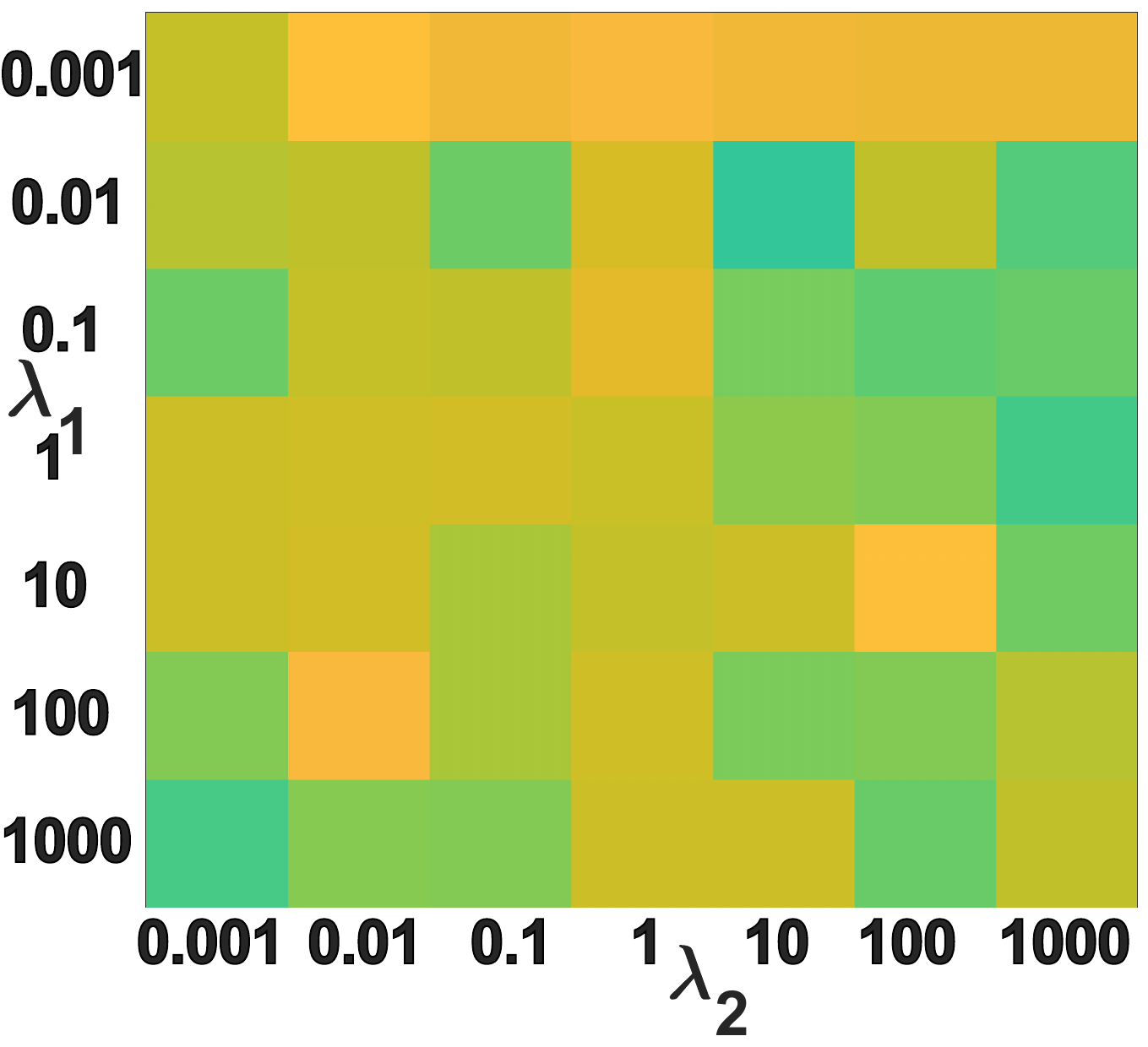}
	}
	\subfigure[COIL40]{
	\includegraphics [height=2.86cm]{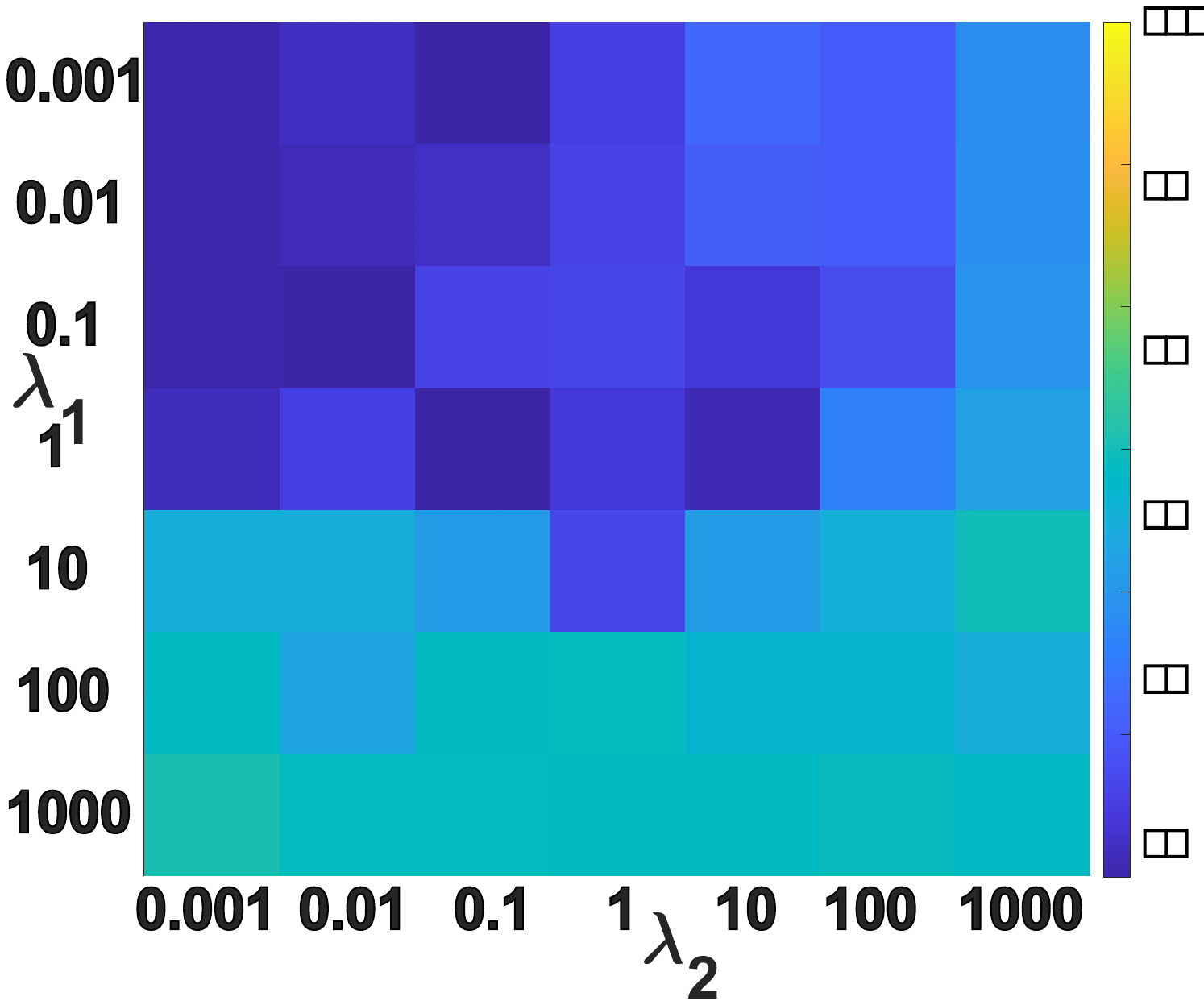}
	}
	\caption{Clustering of the proposed model versus parameters $\lambda_1$ and $\lambda_2$ on five benchmark datasets with the ACC metric.} 
	\label{fig: para_ACC}
\end{figure*}

\begin{figure*}[]
	\centering
	\subfigure[UMIST]{
	\includegraphics [width=3.10cm]{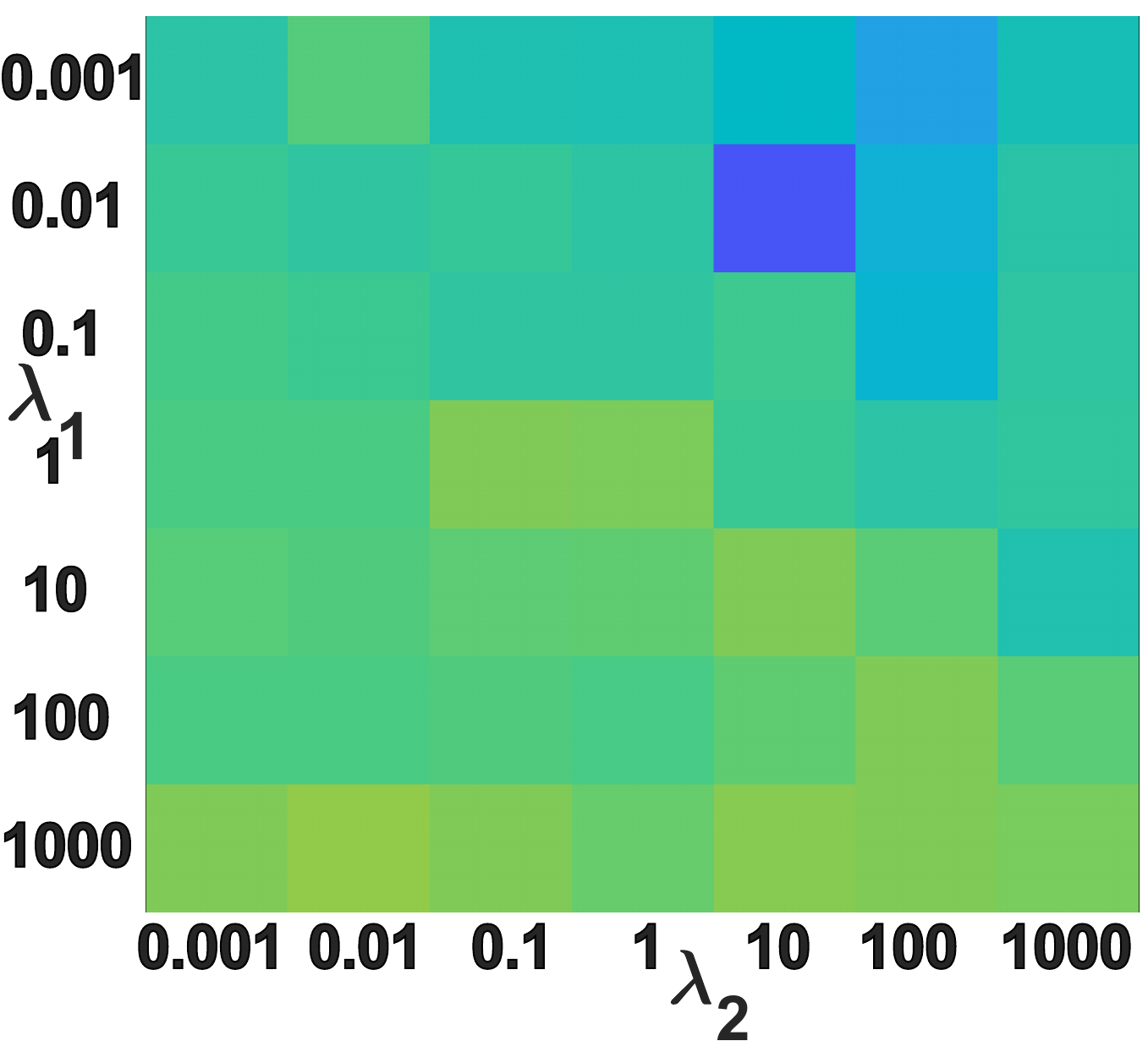}
	}
	\subfigure[ORL]{
	\includegraphics [width=3.10cm]{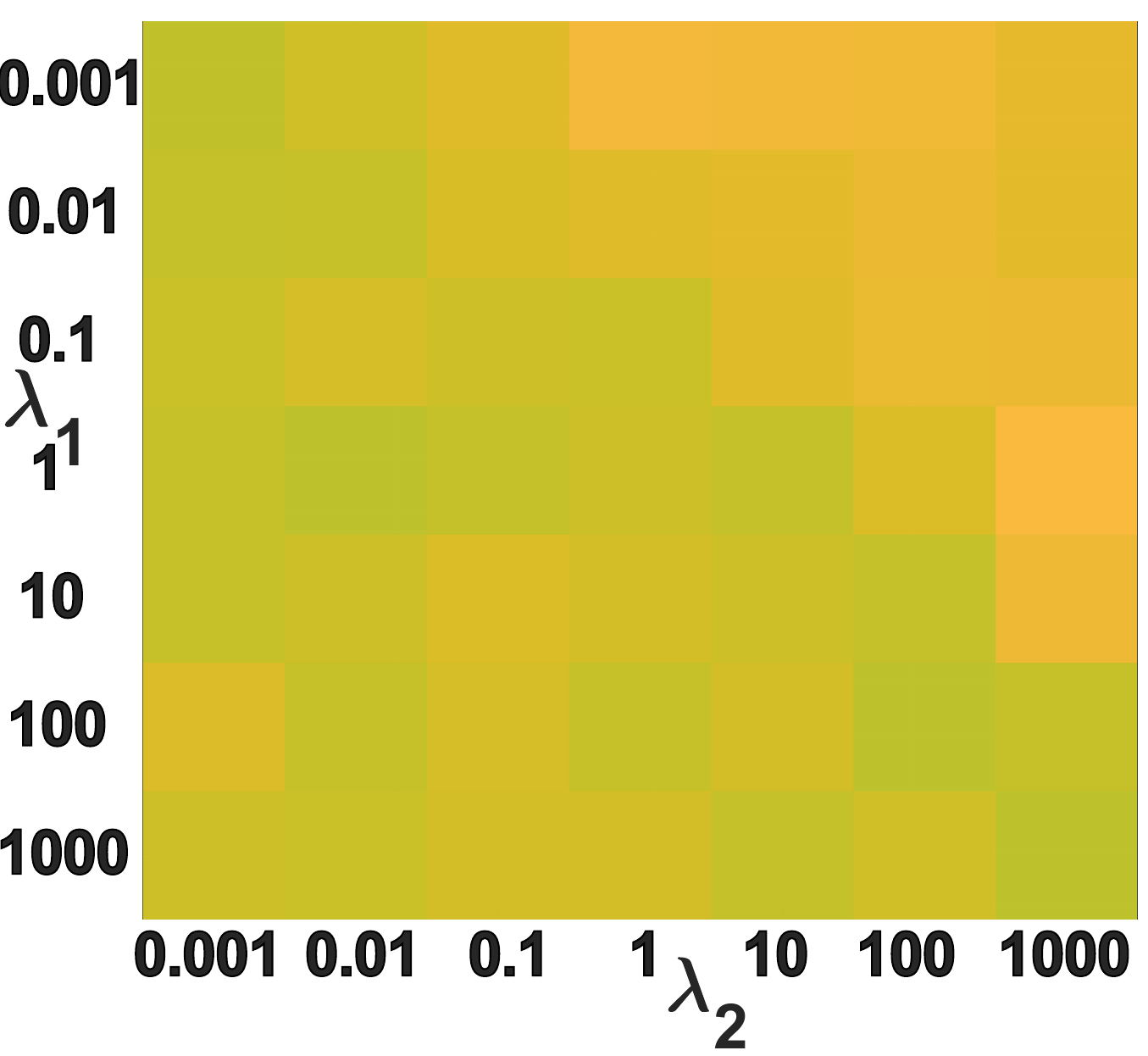}
	}
	\subfigure[MNIST]{
	\includegraphics [width=3.10cm]{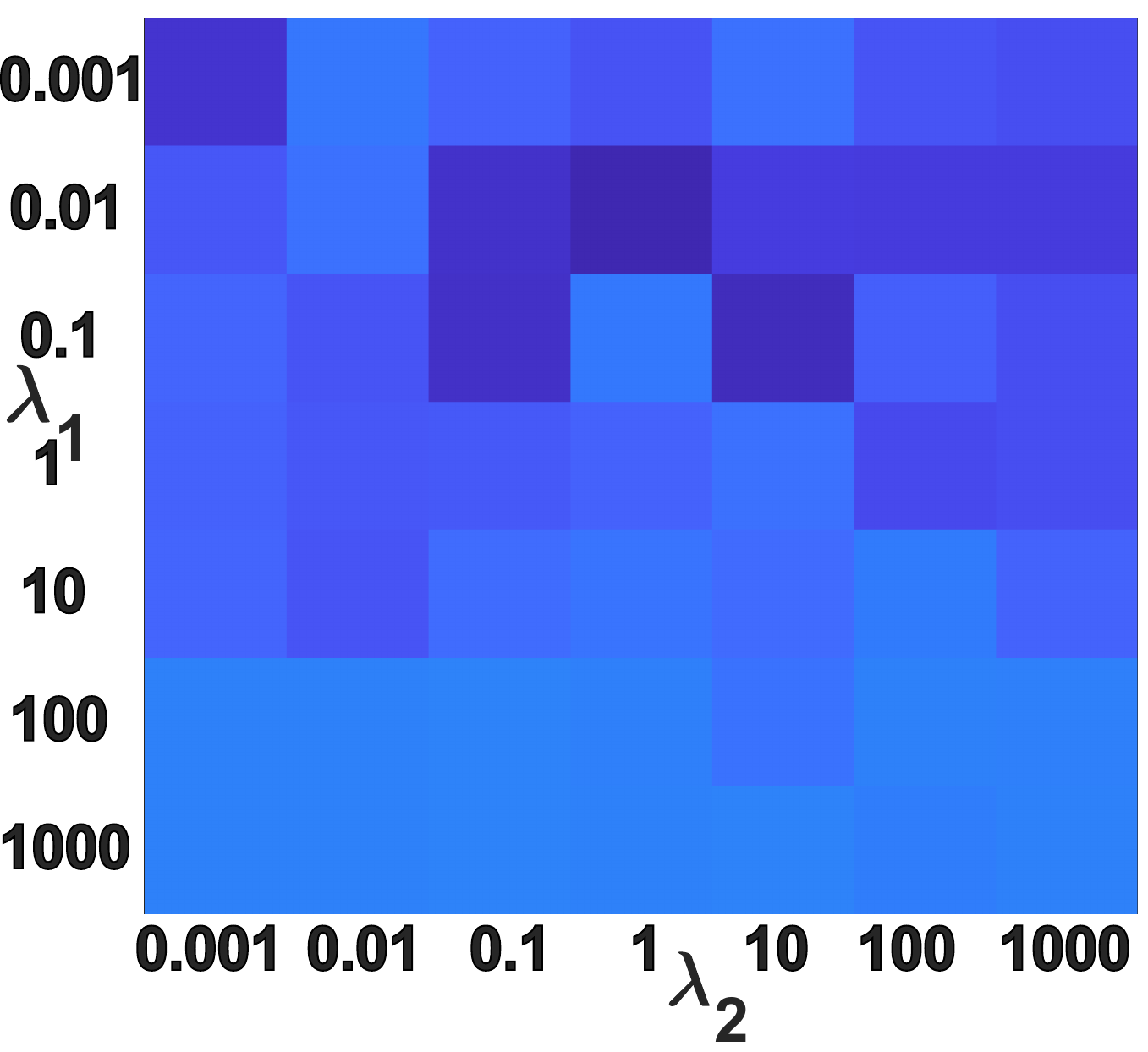}
	}
	\subfigure[COIL20]{
	\includegraphics [width=3.10cm]{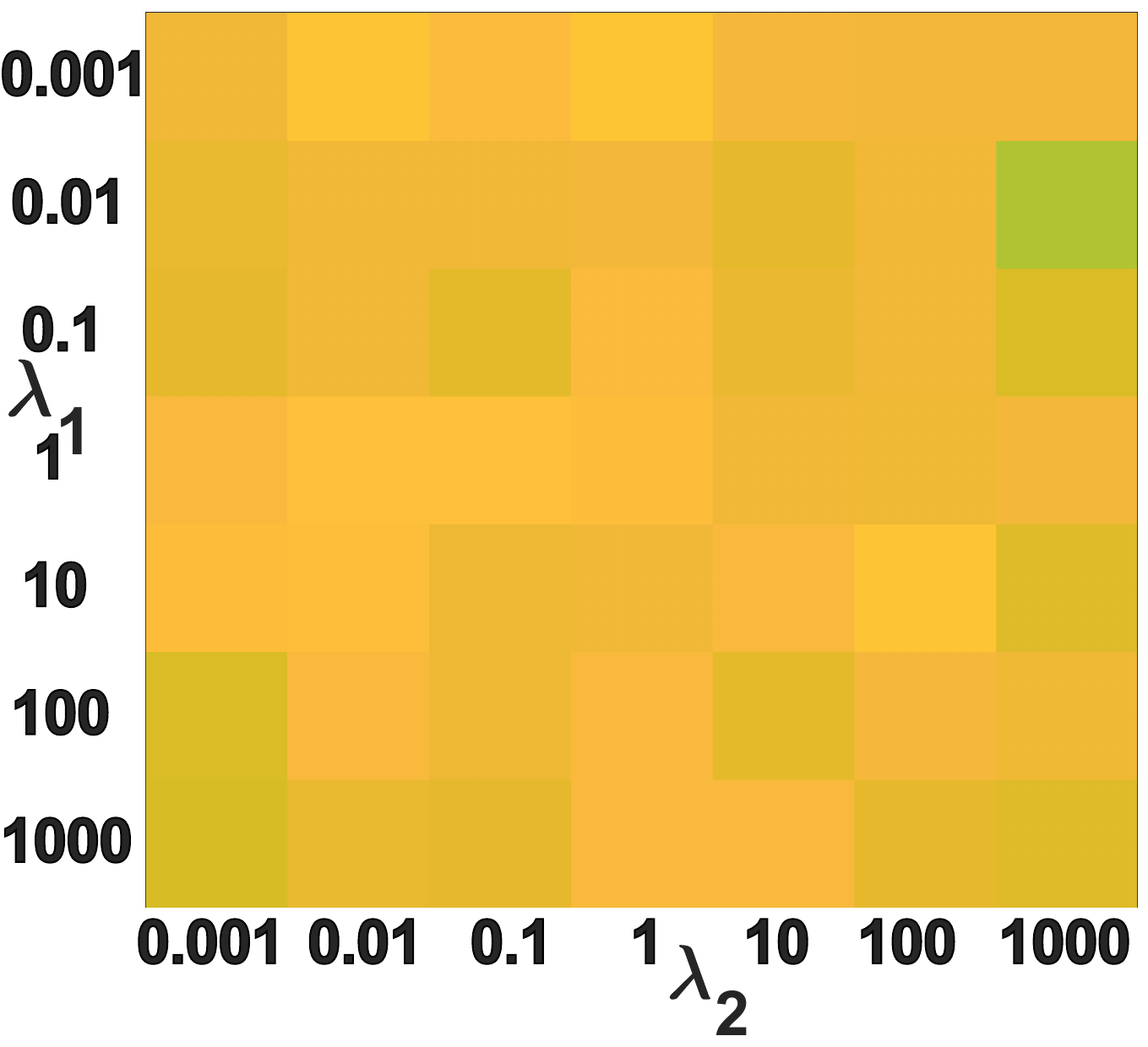}
	}
	\subfigure[COIL40]{
	\includegraphics [height=2.86cm]{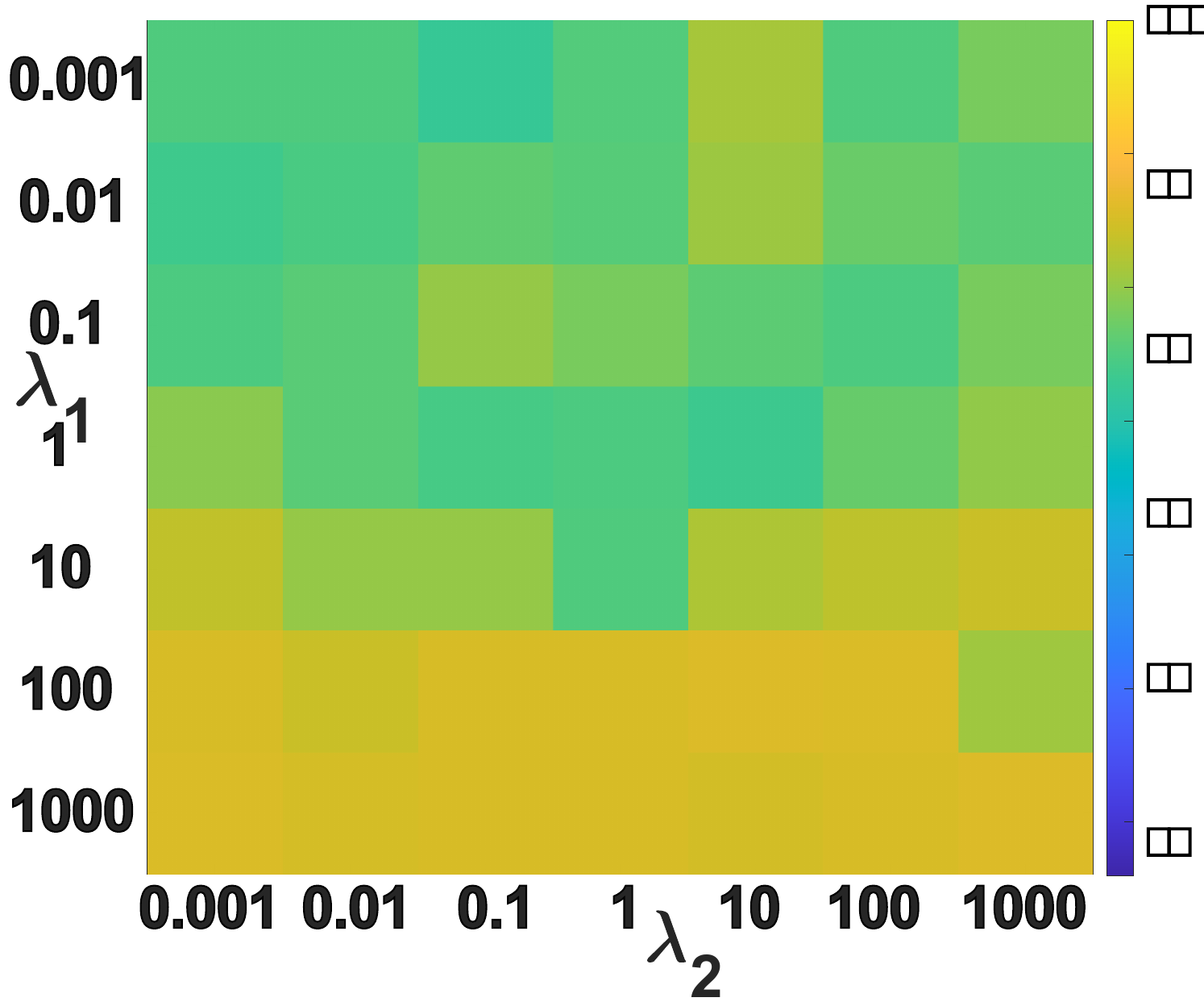}
	}
	\caption{Clustering of the proposed model versus parameters $\lambda_1$ and $\lambda_2$ on five benchmark datasets with the NMI metric.}
	\label{fig: para_NMI}
\end{figure*}

\begin{figure*}[]
	\centering
	\subfigure[UMIST]{
	\includegraphics [width=3.10cm]{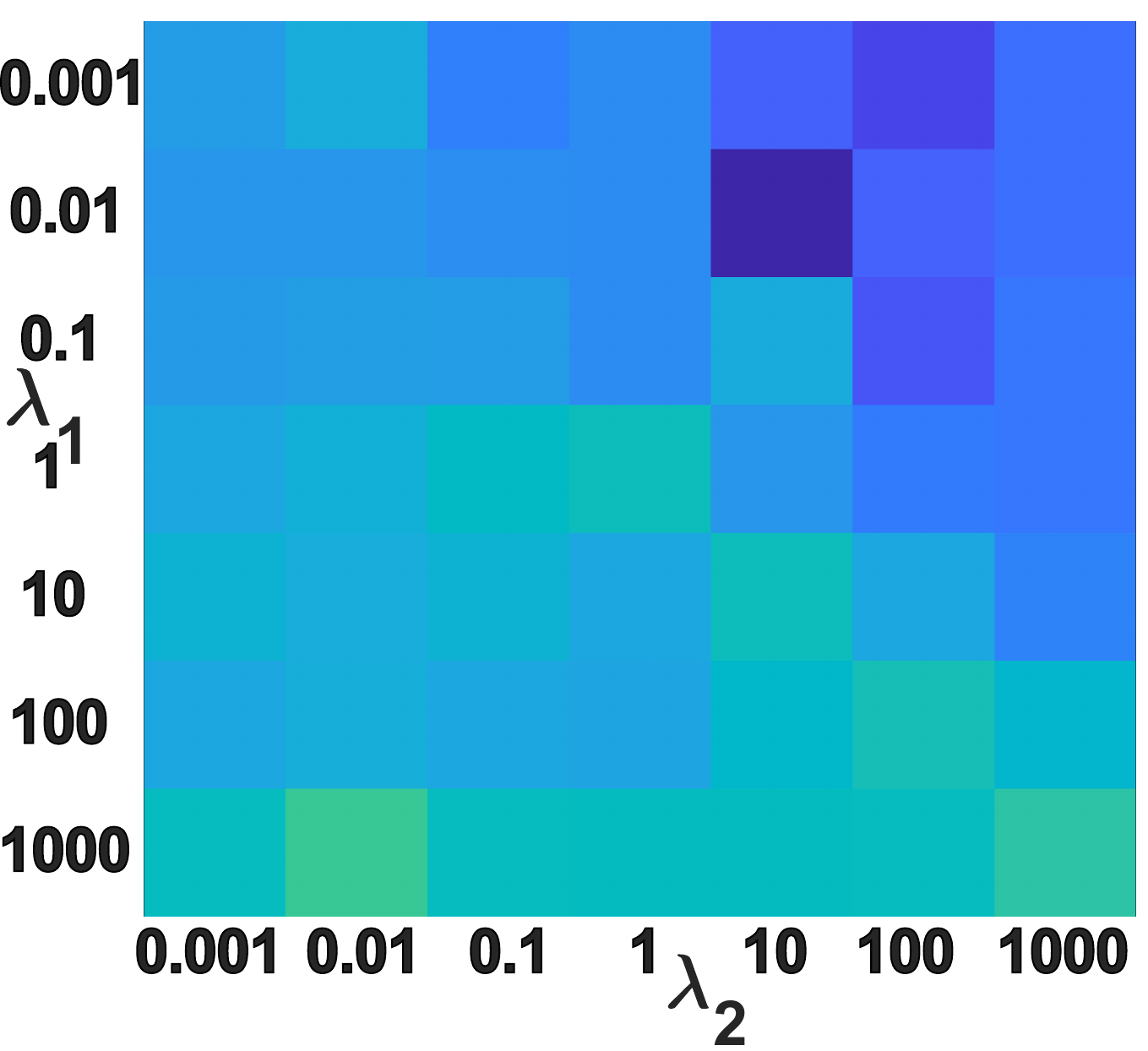}
	}
	\subfigure[ORL]{
	\includegraphics [width=3.10cm]{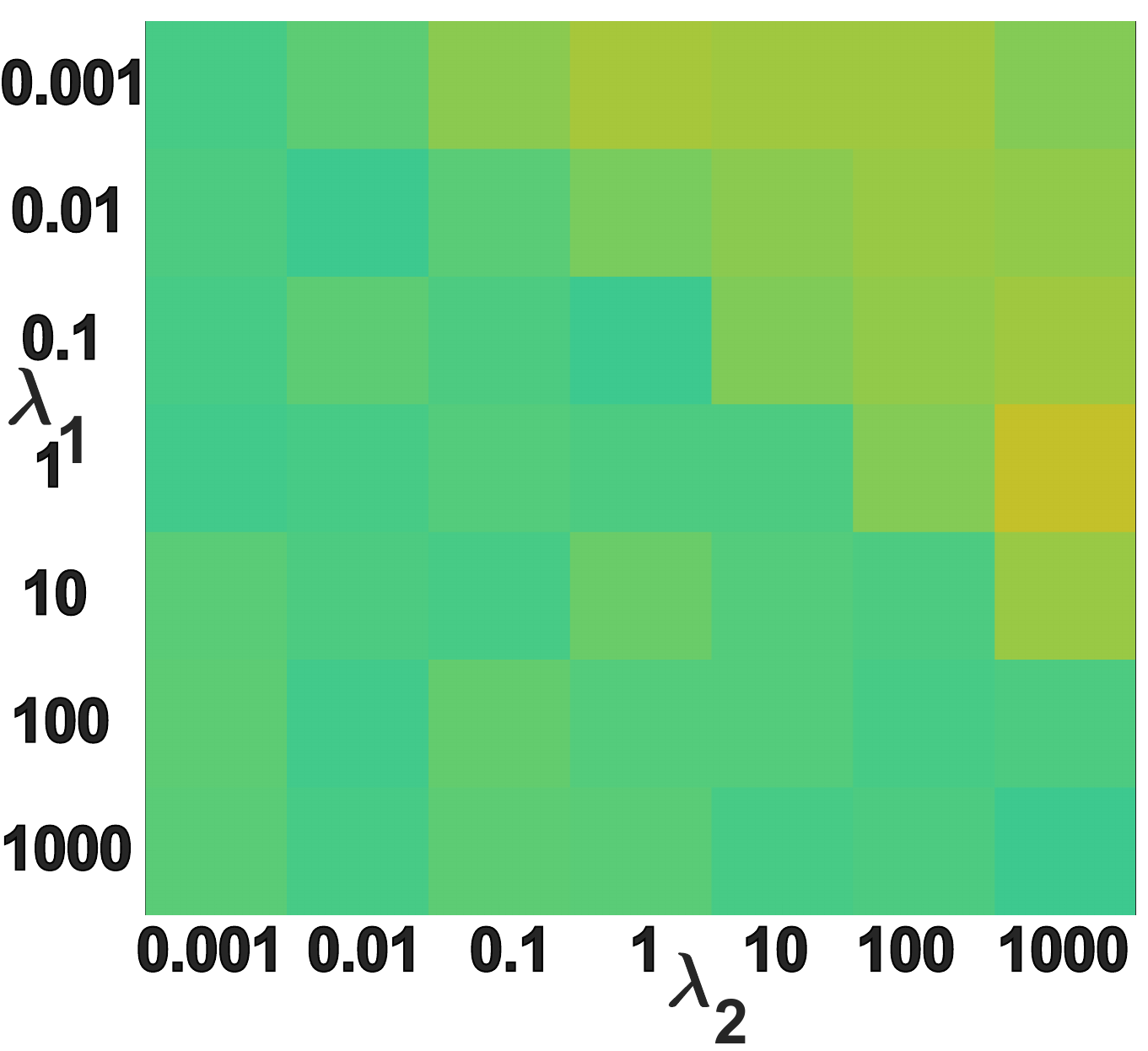}
	}
	\subfigure[MNIST]{
	\includegraphics [width=3.10cm]{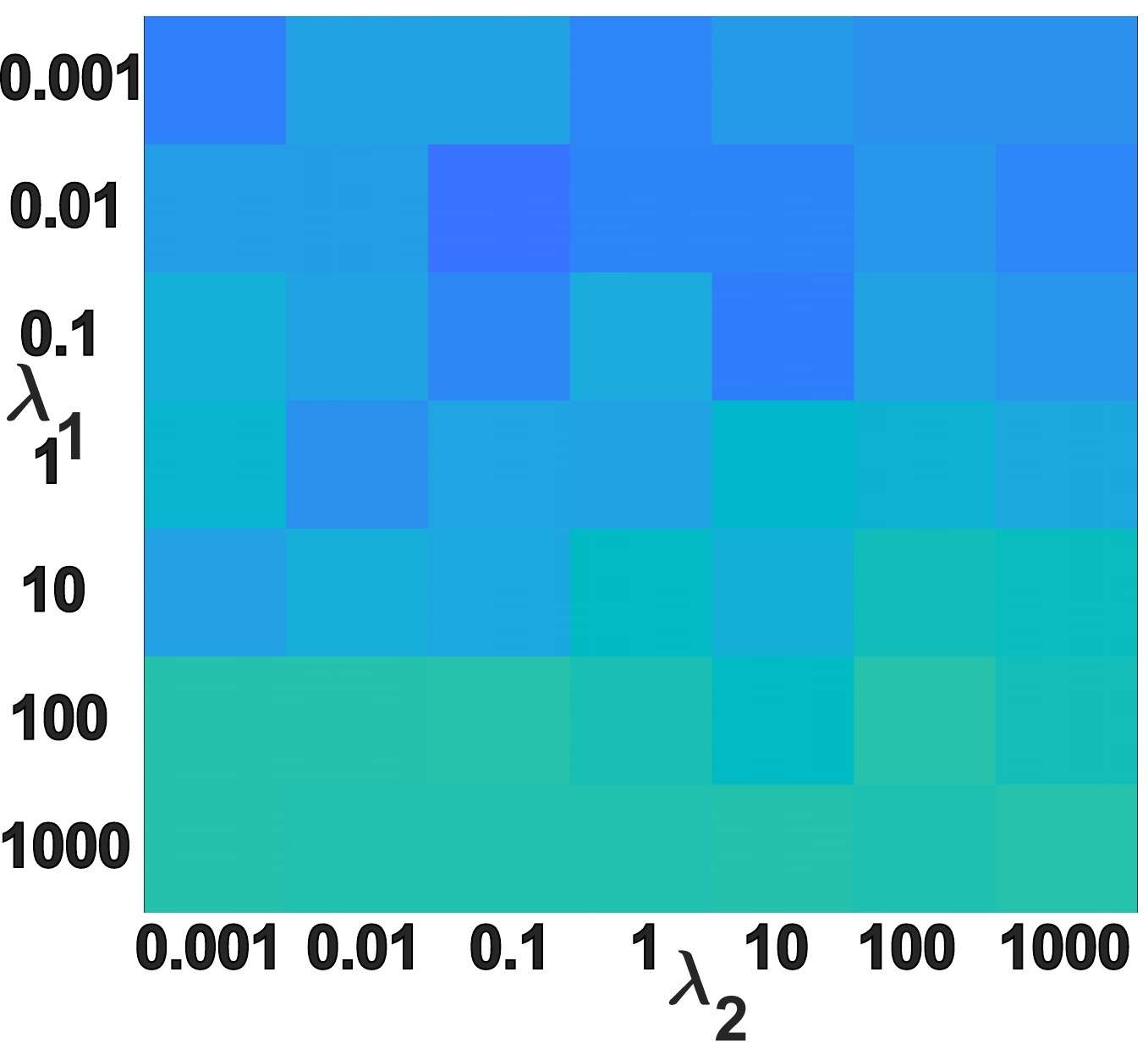}
	}
	\subfigure[COIL20]{
	\includegraphics [width=3.10cm]{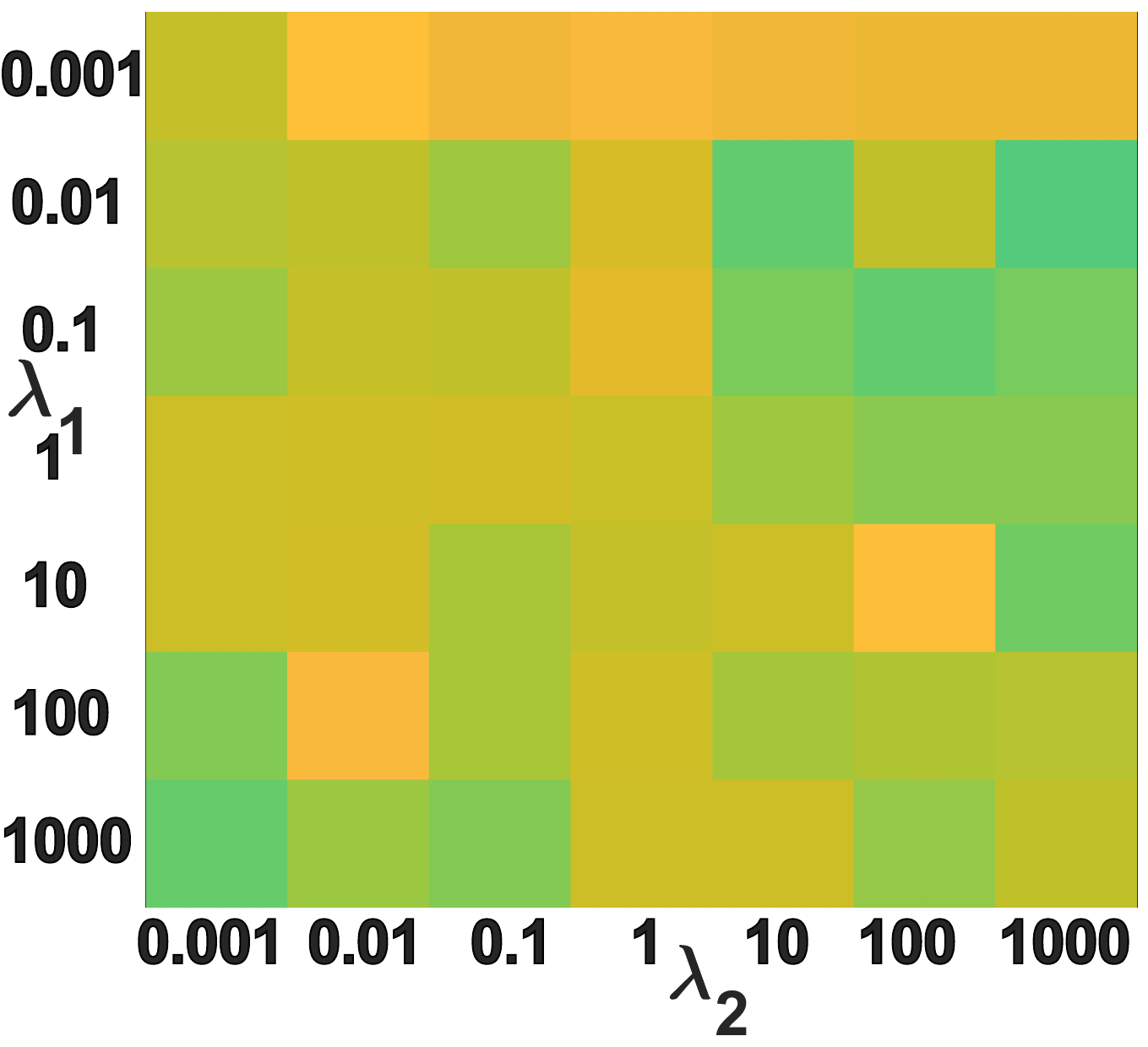}
	}
	\subfigure[COIL40]{
	\includegraphics [height=2.86cm]{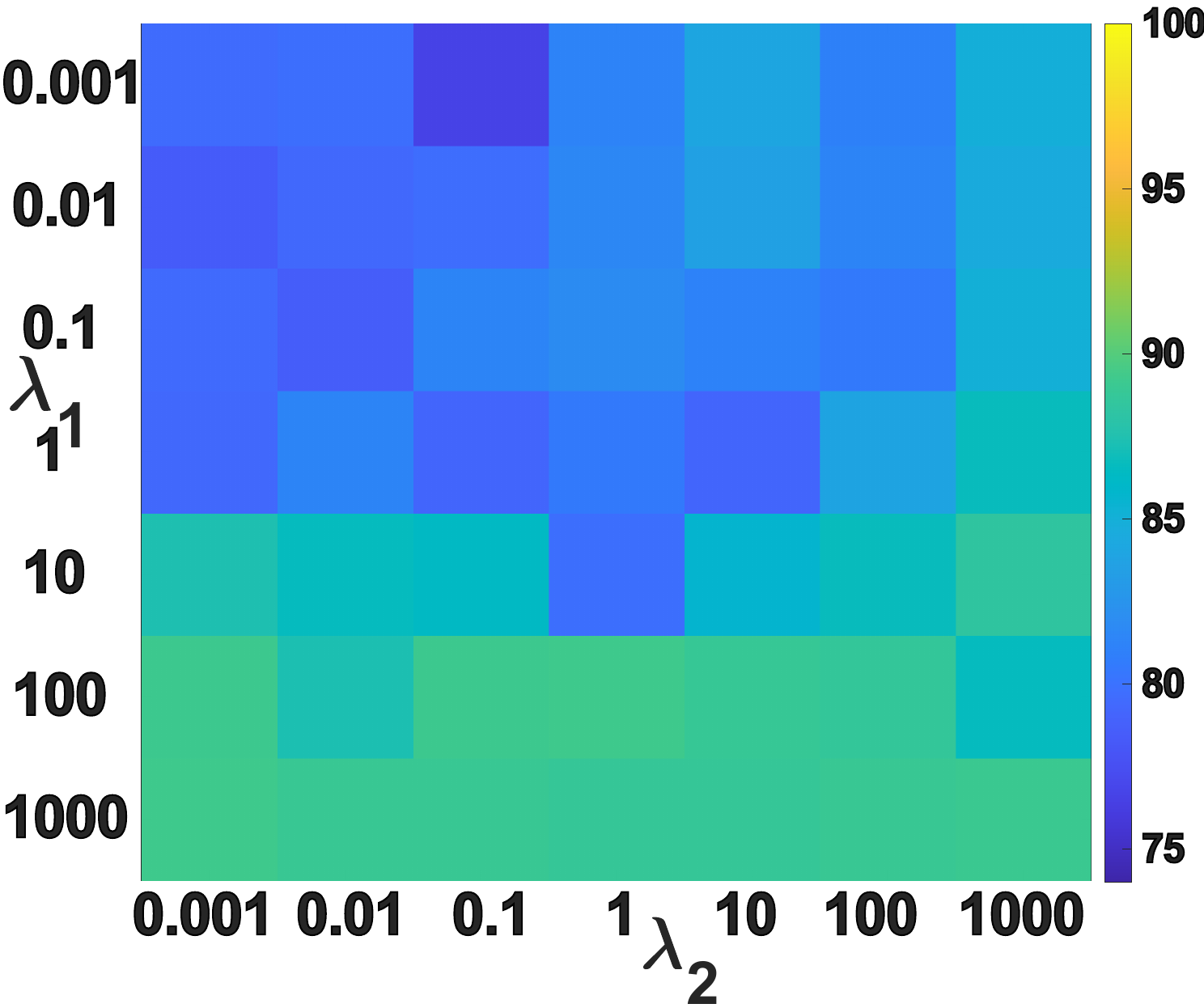}
	}
	\caption{Clustering of the proposed model versus parameters $\lambda_1$ and $\lambda_2$ on five benchmark datasets with the PUR metric.}
	\label{fig: para_PUR}
\end{figure*}

We also visually compared the learned affinity graphs on a synthetic dataset constructed by randomly selecting ten categories from the ORL dataset. As shown in Figure \ref{fig: afm}, it can be seen that the three affinity graphs have an approximate block-diagonal structure; however, $\mathbf{C}_\emph{S}$ and $\mathbf{C}_\emph{F}$ contain fewer error connections than $\mathbf{C}_\emph{A}$. In addition, Figure \ref{fig: cfm} shows the corresponding confusion matrices, in which we can observe that $\mathbf{C}_\emph{F}$ obtains higher clustering accuracy than $\mathbf{C}_\emph{A}$ and $\mathbf{C}_\emph{S}$, benefiting from the adaptive utilization of both the attribute and structure information.

\subsubsection{Structure Matrix Construction} 

In addition to the method for constructing $\mathbf{Z}_\emph{S}$ in Section \ref{subsec: s-se}, we also investigated another three construction manners: 
\begin{table}
\centering
\caption{Investigation of the effect of the manner of constructing the structure matrix $\mathbf{Z}_\emph{S}$ on the ORL dataset.}
\label{tab: structure}
\resizebox{0.42\textwidth}{!}{
\begin{tabular}{c|c|ccc}
\hline\hline
Case & $\mathbf{Z}_\emph{S}$ & ACC                     & NMI                     & PUR \\ 
\hline\hline
\multirow{2}{*}{1} & \multirow{2}{*}{$\frac{\mathbf{Z}^\mathsf{T}\mathbf{Z}}{\left\|\mathbf{Z}^\mathsf{T}\right\|_F\left\|\mathbf{Z}\right\|_F}$}     
& \multirow{2}{*}{0.8675} & \multirow{2}{*}{0.9177} & \multirow{2}{*}{0.8750} \\
                      &                         &                         &    \\ \hline
\multirow{2}{*}{2} & \multirow{2}{*}{$\mathbf{C}_\emph{A}$}     
& \multirow{2}{*}{0.8925} & \multirow{2}{*}{0.9357} & \multirow{2}{*}{0.9000} \\
                      &                         &                         &                       \\ \hline
\multirow{2}{*}{3} & \multirow{2}{*}{$\frac{|\mathbf{C}_\emph{A}|+|\mathbf{C}_\emph{A}^\mathsf{T}|}{2}$}     
& \multirow{2}{*}{0.7450} & \multirow{2}{*}{0.8818} & \multirow{2}{*}{0.7975} \\

                      &                         &                         &    \\ \hline   
\multirow{2}{*}{4} & \multirow{2}{*}{$\frac{\mathbf{C}_\emph{A}+\mathbf{C}_\emph{A}^\mathsf{T}}{2}$}     
& \multirow{2}{*}{\textbf{}} & \multirow{2}{*}{\textbf{}} & \multirow{2}{*}{\textbf{}} \\
                      &  & \textbf{0.9075}    & \textbf{0.9431}    & \textbf{0.9175}         \\   
\hline\hline  
\end{tabular}
}
\end{table}

\begin{itemize}
    \item We constructed $\mathbf{Z}_\emph{S}$ by using the cosine similarity, i.e.,
    \begin{equation}
        \begin{aligned}
        \mathbf{Z}_\emph{S}=\frac{\mathbf{Z}^\mathsf{T}\mathbf{Z}}{\left\|\mathbf{Z}^\mathsf{T}\right\|_F\left\|\mathbf{Z}\right\|_F}.
        \label{eq: cosine}
        \end{aligned}
    \end{equation}
    \item We directly utilized $\mathbf{C}_\emph{A}$ as the structure matrix:
    \begin{equation}
        \begin{aligned}
        \mathbf{Z}_\emph{S}=\mathbf{C}_\emph{A}.
        \label{eq: cv2ze}
        \end{aligned}
    \end{equation}
    \item We further enforced $\mathbf{Z}_\emph{S}$ to satisfy the symmetric and non-negative properties, i.e.,
    \begin{equation}
        \begin{aligned}
        \mathbf{Z}_\emph{S}=\frac{|\mathbf{C}_\emph{A}|+|\mathbf{C}_\emph{A}^\mathsf{T}|}{2}.
        \label{eq: symm}
        \end{aligned}
    \end{equation}
\end{itemize}

Table \ref{tab: structure} lists the corresponding results of different construction methods on the ORL dataset, where we have the following observations.
\begin{itemize}
    \item \textit{Case} $4$ outperforms the others in terms of all the metrics considerably. For example, it improves the ACC, NMI, and PUR of the second-best way (i.e., \textit{Case} $2$) by 1.50\%, 0.74\%, and 1.75\%, respectively. In particular, the comparison with \textit{Case} $3$ validates the effectiveness of preserving the negative values of $\mathbf{C}_\emph{A}$ as the negative values can also indicate the discriminative information among samples. 
    \item \textit{Case} $1$ performs poorly compared with \textit{Cases} $2$ and $4$. This phenomenon suggests that the cosine similarity is not a good choice for constructing the structure matrix in our framework. The reason may be that the cosine similarity only considers the angle between two vectors while ignoring the magnitude, which is important to indicate the similarity of two samples.
\end{itemize}

\begin{figure*}[]
	\centering
	\subfigure[UMIST]{
	\includegraphics [width=3.2cm]{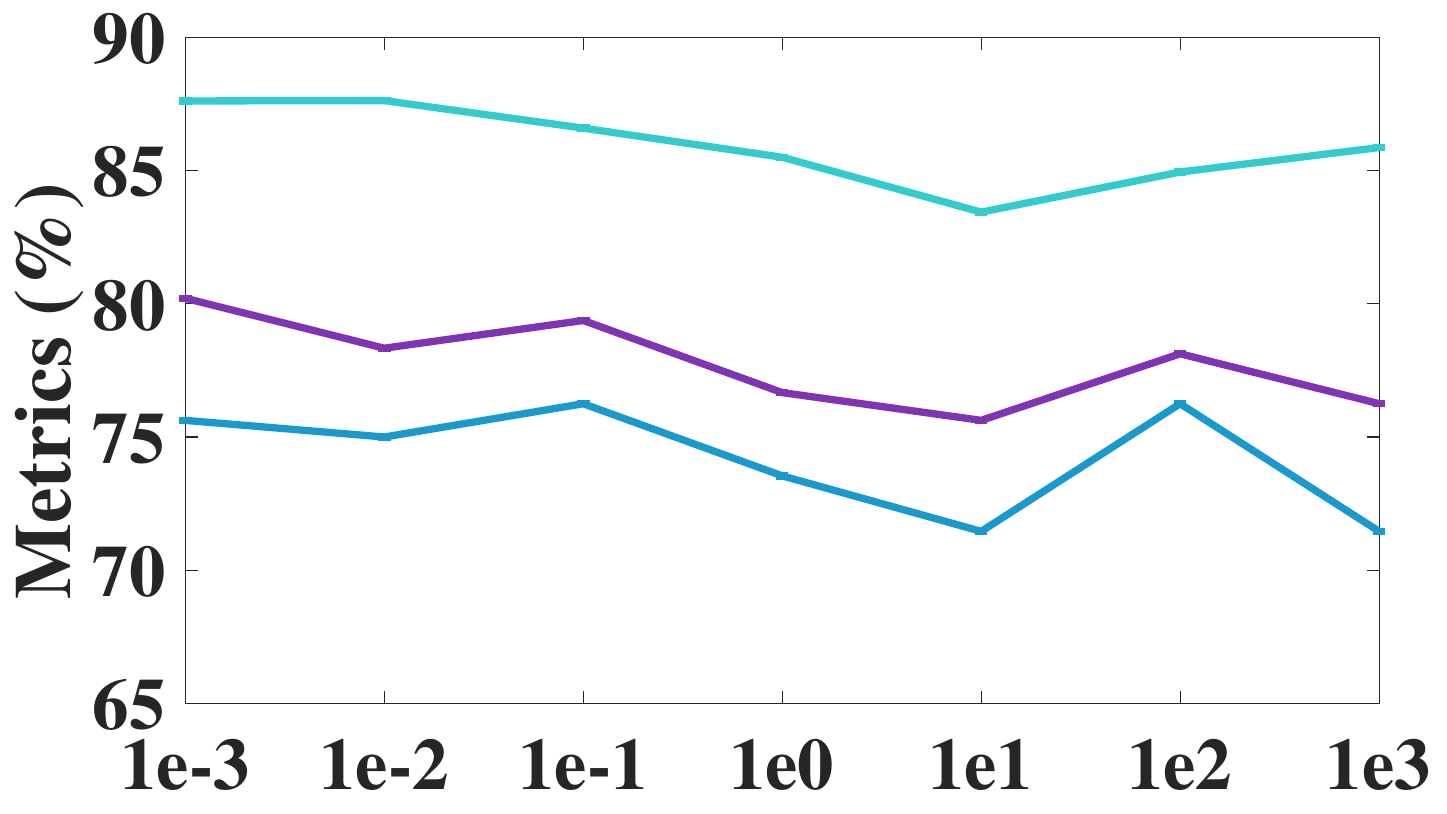}
	}
	\subfigure[ORL]{
	\includegraphics [width=3.2cm]{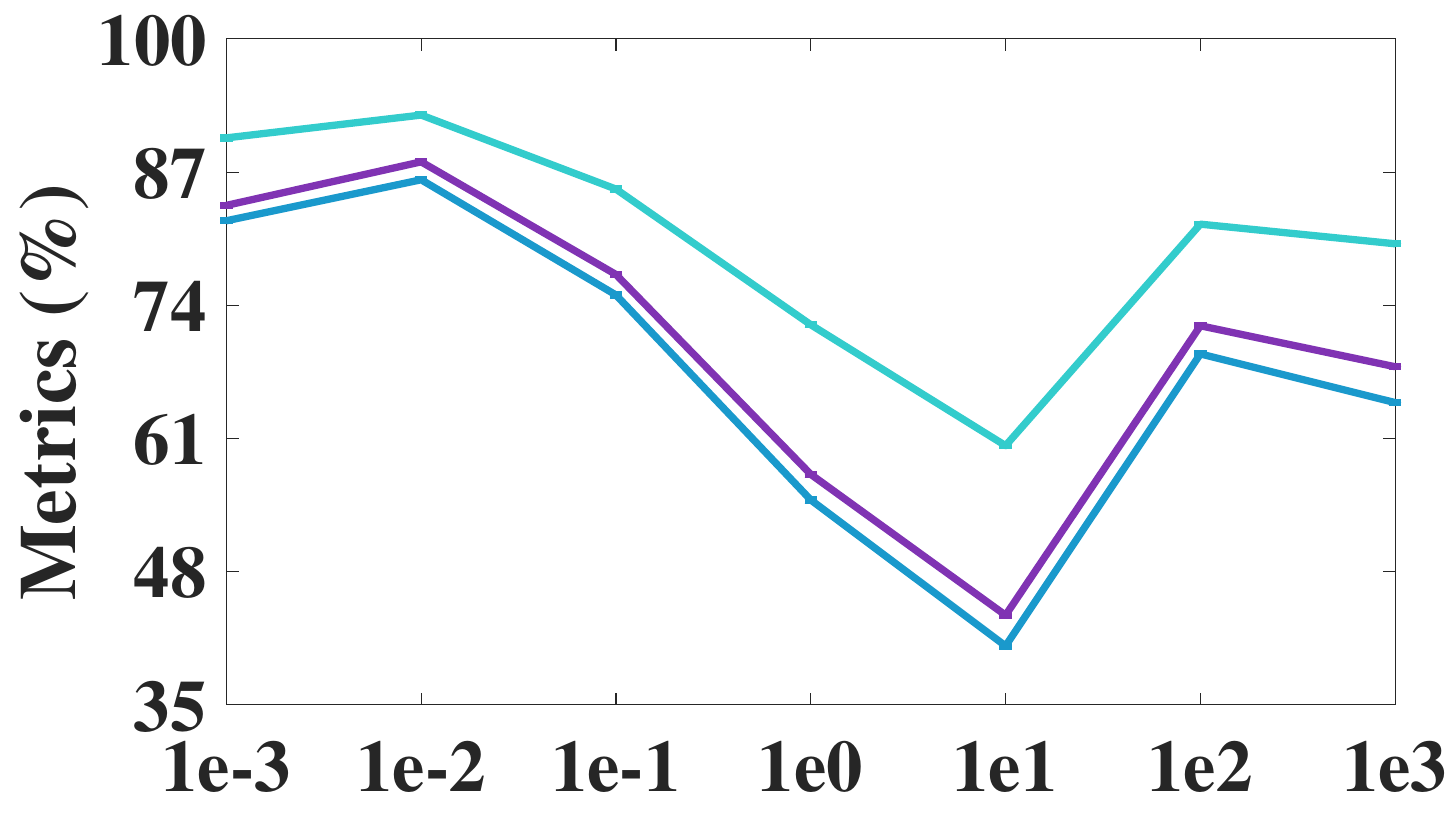}
	}
	\subfigure[MNIST]{
	\includegraphics [width=3.2cm]{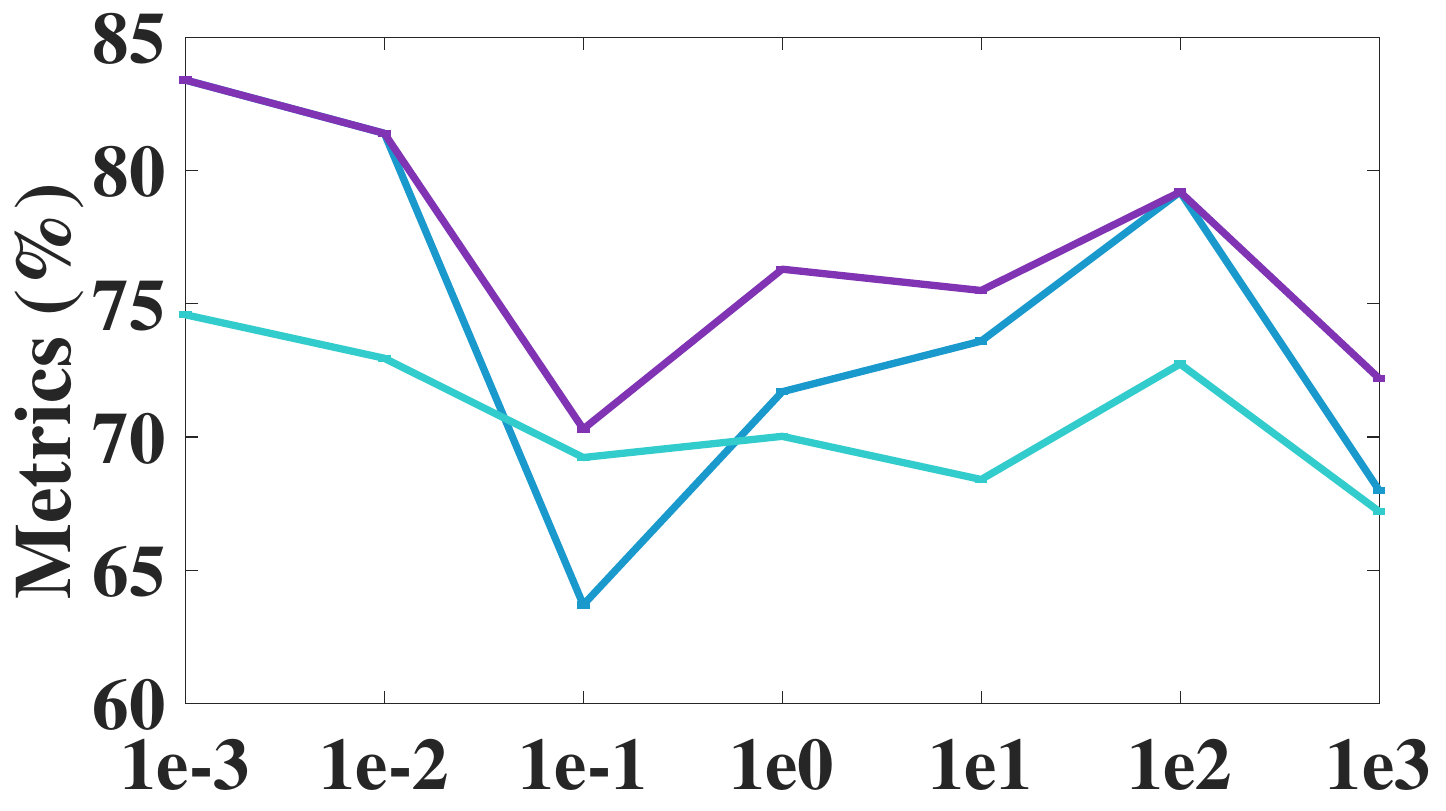}
	}
	\subfigure[COIL20]{
	\includegraphics [width=3.2cm]{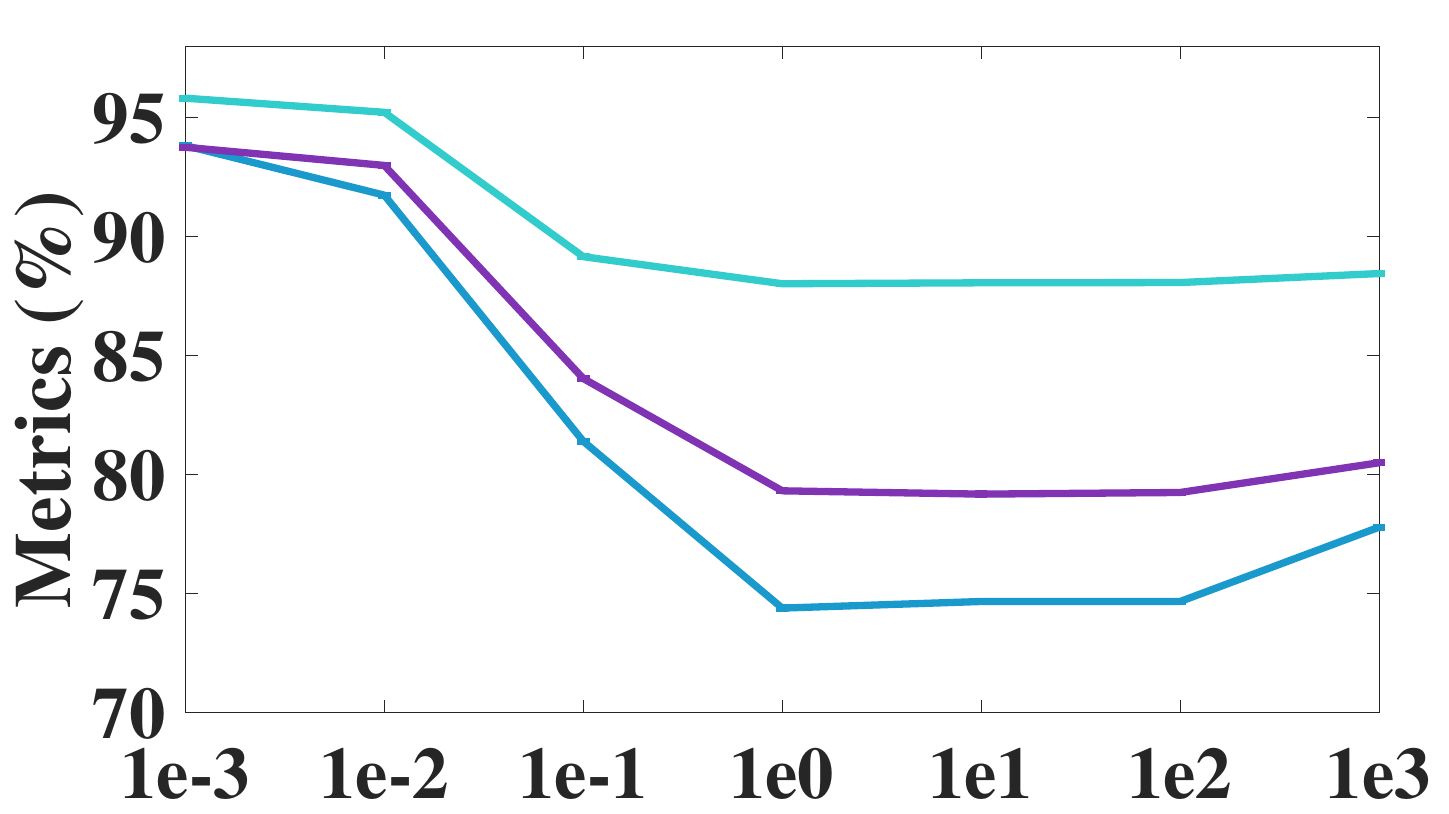}
	}
	\subfigure[COIL40]{
	\includegraphics [width=3.2cm]{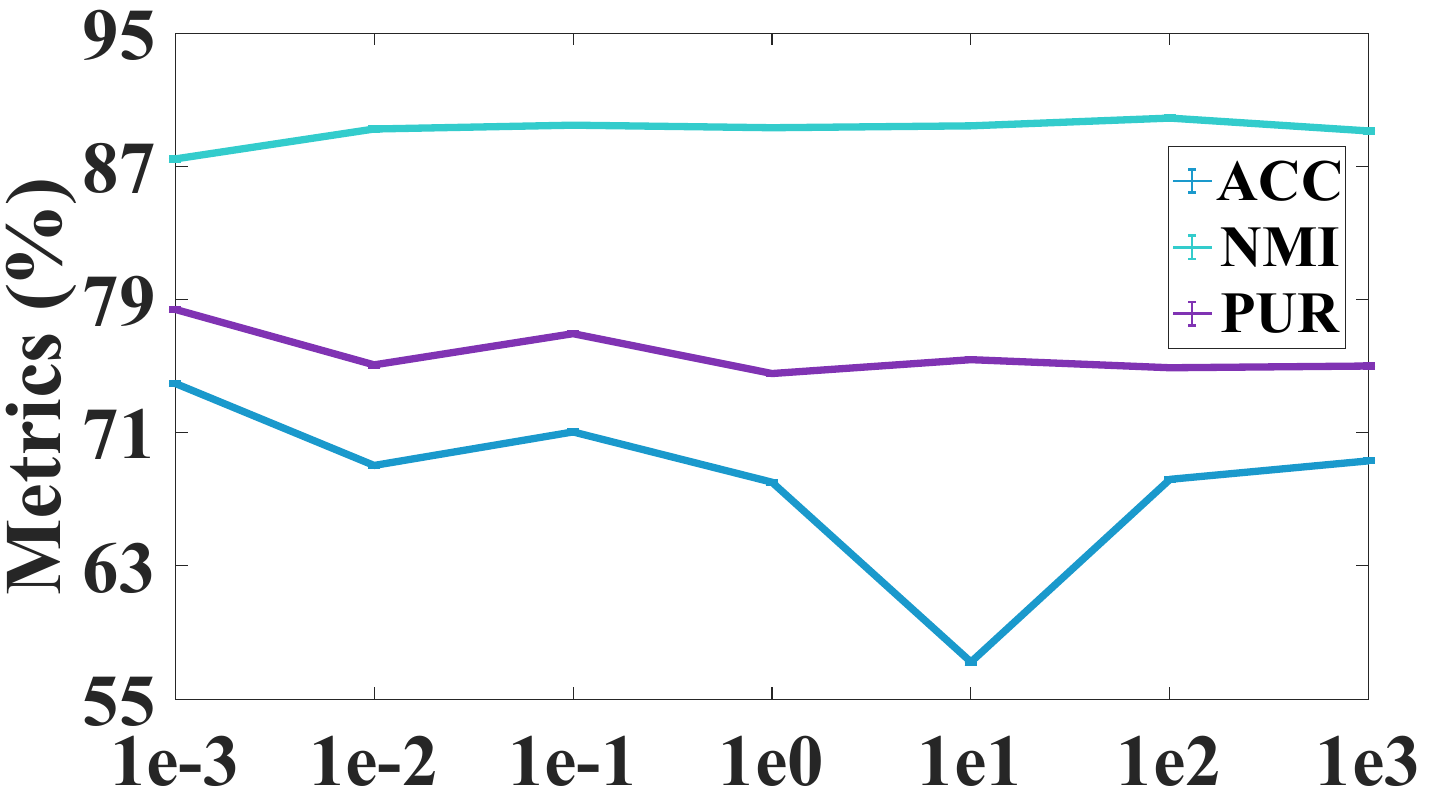}
	}
	\caption{Clustering of the proposed model versus parameters $\beta$ on benchmark datasets with three metrics.}
	\label{fig: para_LD3S}
\end{figure*}

\subsection{Visual Comparison}
We used the 2D t-distributed stochastic neighbor embedding (t-SNE) technique \cite{maaten2008visualizing} to visualize the encoder output representations of our method and two representative deep clustering methods on one generated dataset constructed by randomly selecting ten categories from COIL20. From Figure \ref{fig:tsne}, we can observe that the produced embedding by our method has the best accuracy and separability for different clusters, indicating that our AASSC-Net could learn a more discriminative representation than the compared methods.

\subsection{Parameter Analysis}
In the loss function Eq. (\ref{eq: Our}), there are four hyperparameters (i.e., $\gamma$, $\lambda_1$, $\lambda_2$, and $\beta$) used to balance the contributions of the attribute SE, the structure affinity graph, the structure SE, and the attention-based fusion respectively. We first set the same $\gamma$ as the optimal parameter of \cite{ji2017deep}. Then, we selected $\lambda_1$ and $\lambda_2$ from the set $\mathcal{S}=\left\{ 0.001,0.01,0.1,1,10,100,1000 \right\}$ to evaluate their influence. Figures \ref{fig: para_ACC}, \ref{fig: para_NMI} and \ref{fig: para_PUR} show the values of ACC, NMI, and PUR w.r.t. different $\lambda_1$ and $\lambda_2$. \textcolor{black}{We finally tuned $\beta$ from the set $\mathcal{S}$ by fixing other parameters and the results are shown in Figure \ref{fig: para_LD3S}. From those figures, we have the following observations.}
\begin{itemize}
    \item \textcolor{black}{On UMIST, the relatively small $\lambda_1$ and large $\lambda_2$ lead to low values of ACC, NMI, and PUR. The reason may be that the UMIST dataset includes similar face images, and thus when conducting the SE, the regularization term of learning the structure affinity matrix is vital for keeping the connectivity within the same subspace.} 
    \item \textcolor{black}{On ORL, MNIST, and COIL20, our method is capable of achieving almost optimal performance in terms of the three metrics in a wide range of $\lambda_1$ and $\lambda_2$, illustrating that our method is robust to $\lambda_1$ and $\lambda_2$.}
    \item \textcolor{black}{On COIL40, the relatively small values of $\lambda_1$ and $\lambda_2$ produce an apparent performance degradation on all metrics. Such a phenomenon indicates the importance of $\lambda_1$ and $\lambda_2$.}
    \item \textcolor{black}{The proposed model is relatively sensitive to $\beta$. Generally, a relatively small $\beta$ can significantly improve the clustering performance, while a relatively large one will harm the clustering performance.}
\end{itemize}

\section{Conclusion}
\label{sec: conclusion}
We presented a novel deep self-expressiveness-based subspace clustering framework. Specifically, we first constructed the attribute and structure matrices separately after nonlinearly projecting the raw data into the latent feature space. Then, we employed the SE priori to learn affinity graphs from the constructed attribute and structure matrices separately. Afterward, we exploited an attention-based strategy to fuse the two learned affinity graphs. We quantitatively and qualitatively validated the advantage of the proposed method over eighteen state-of-the-art methods on five commonly-used benchmark datasets. In addition, we performed comprehensive ablation studies to validate the effectiveness of the designed modules. \textcolor{black}{In the future, we will investigate the potential of our method on large-scale datasets by using the parametric approximation of the self-expressiveness matrices \cite{zhang2021learning}.}

\balance
\bibliographystyle{IEEEtran}
\bibliography{AASSC}

\begin{IEEEbiography}[{\includegraphics[width=1in,height=1.25in,clip,keepaspectratio]{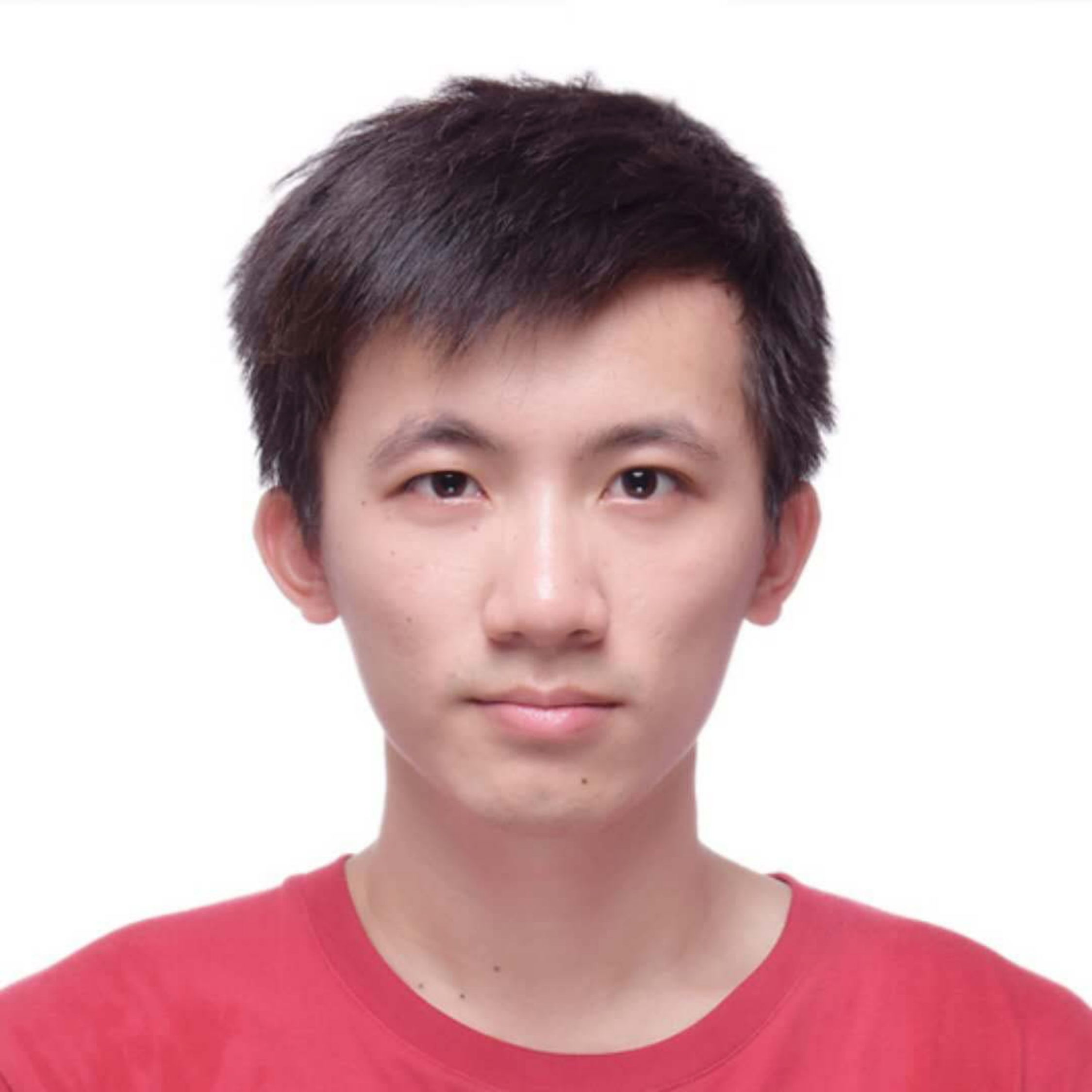}}]{Zhihao Peng}
received the B.S. and M.S. degrees in computer science and technology from Guangdong University of Technology, Guangzhou, China, in 2016 and 2019, respectively. He is currently pursuing the Ph.D. degree in department of computer science from City University of Hong Kong, SAR, China.

His current research interests include unsupervised data modeling, subspace clustering, and machine learning with graphs.
\end{IEEEbiography}

\begin{IEEEbiography}[
 {
  \includegraphics[width=1in,height=1.25in,clip,keepaspectratio]{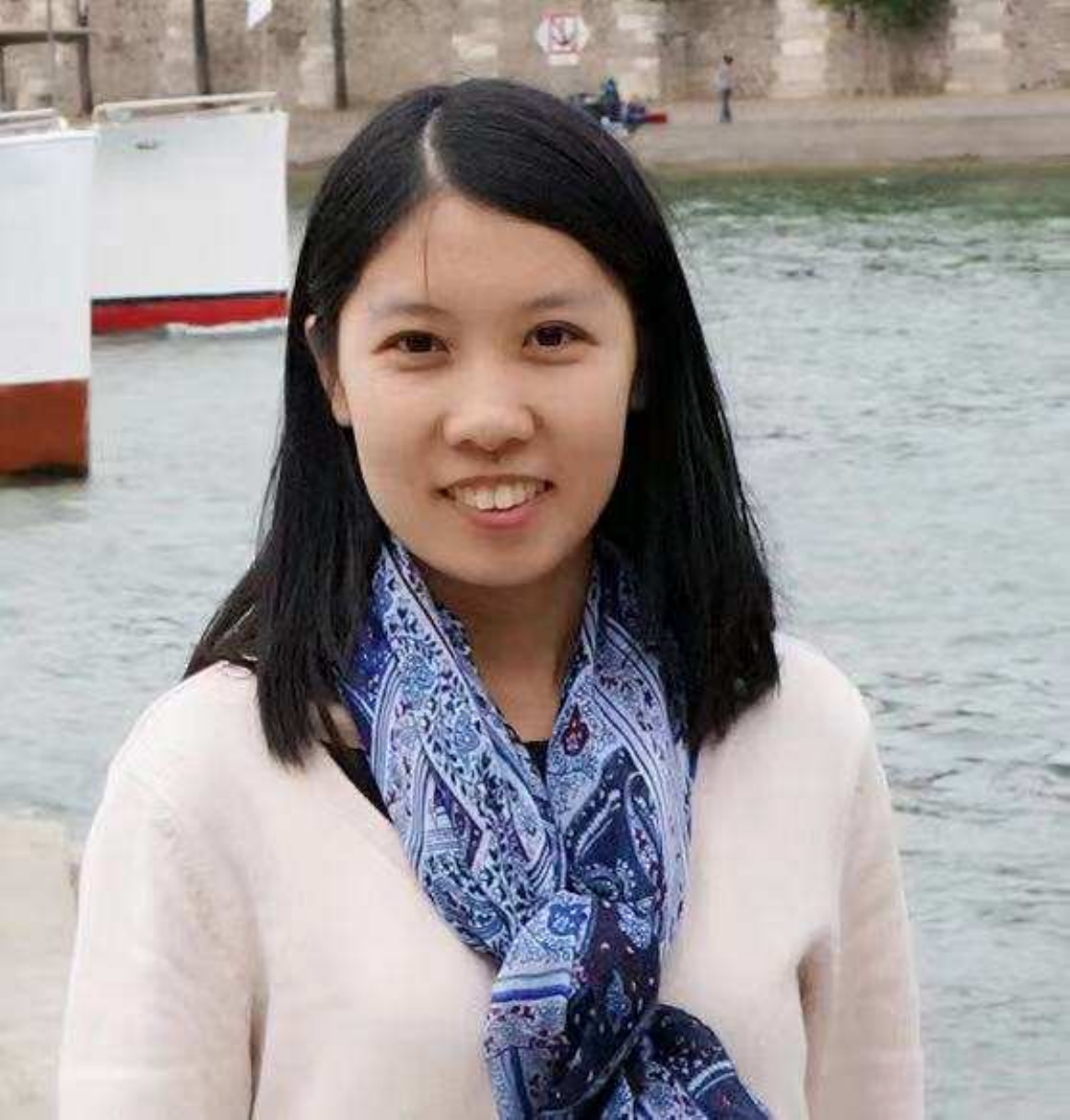}
 }
 ]{Hui Liu}
received the B.Sc. degree in Communication Engineer from the Central South University, Changsha, China, the M.Eng. degree in Computer Science from Nanyang Technological University, Singapore, and the Ph.D. degree from the Department of Computer Science, City University of Hong Kong, Hong Kong. 

She is currently a Lecturer with the School of Computing Information Sciences, Caritas Institute of Higher Education, Hong Kong. From 2014 to 2017, she was a research associate at the Maritime Institute of Nanyang Technological University. Her research interests include image processing and machine learning.

\end{IEEEbiography}

\begin{IEEEbiography}[
 {
  \includegraphics[width=1in,height=1.25in,clip,keepaspectratio]{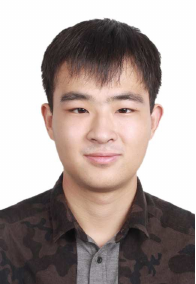}
 }
 ]{Yuheng Jia}
received the B.S. degree in automation and the M.S. degree in control theory and engineering from Zhengzhou University, Zhengzhou, China, in 2012 and 2015, respectively, and the Ph.D. degree in computer science from the City University of Hong Kong, SAR, China, in 2019. 

He is currently an associate professor with the School of Computer Science and Engineering, Southeast University, China. 
His research interests include machine learning, Bayesian method, spectral clustering and low-rank modeling.
\end{IEEEbiography}

\begin{IEEEbiography}[{\includegraphics[width=1in,height=1.25in,clip,keepaspectratio]{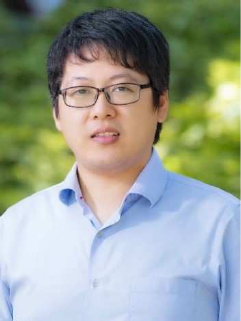}}]{Junhui Hou}
(Senior Member) is an Assistant Professor with the Department of Computer Science, City University of Hong Kong since Jan. 2017. He received the B.Eng. degree in information engineering (Talented Students Program) from the South China University of Technology, Guangzhou, China, in 2009, the M.Eng. degree in signal and information processing from Northwestern Polytechnical University, Xian, China, in 2012, and the Ph.D. degree in electrical and electronic engineering from the School of Electrical and Electronic Engineering, Nanyang Technological University, Singapore, in 2016. His research interests fall into the general areas of multimedia signal processing, such as image/video/3D geometry data representation, processing and analysis, semi/un-supervised data modeling, and data compression.

He received the Chinese Government Award for Outstanding Students Study Abroad from China Scholarship Council in 2015 and the Early Career Award (3/381) from the Hong Kong Research Grants Council in 2018. He is an elected member of MSA-TC and VSPC-TC, IEEE CAS, and MMSP-TC, IEEE SPS. He is currently an Associate Editor for IEEE Transactions on Image Processing, IEEE Transactions on Circuits and Systems for Video Technology, Signal Processing: Image Communication, and The Visual Computer. He also served as the Guest Editor for the IEEE Journal of Selected Topics in Applied Earth Observations and Remote Sensing and as an Area Chair of ACM MM’19/20/21/22, IEEE ICME’20, VCIP’20/21, and WACV’21.
\end{IEEEbiography}

\end{document}